\documentclass[acmtog,natbib=true,nonacm]{acmart}
\usepackage{booktabs} 

\usepackage{times}
\usepackage{epsfig}
\usepackage{amsfonts}
\usepackage{amsmath}
\usepackage{graphicx}
\usepackage{bbding}
\usepackage{tikz}
\usepackage{makecell}
\usepackage{booktabs} 
\usepackage{multirow}
\definecolor{LightCyan}{rgb}{0.935,1,1}

\usepackage[ruled]{algorithm2e}
\SetKwInput{KwInput}{Input}                
\SetKwInput{KwReturn}{Return}
\SetKwInput{KwIntialize}{Initialize} 

\SetAlFnt{\small}
\SetAlCapFnt{\small}
\SetAlCapNameFnt{\small}
\SetAlCapHSkip{0pt}

\newcommand\blfootnote[1]{%
  \begingroup
  \renewcommand\thefootnote{}\footnote{#1}%
  \addtocounter{footnote}{-1}%
  \endgroup
}

\definecolor{red}{rgb}{1,0.6,0.6}
\definecolor{orange}{rgb}{1,0.8,0.6}
\definecolor{yellow}{rgb}{1,1,0.6}

\usepackage{colortbl}

\usepackage{xcolor}

\definecolor{myred}{rgb}{0.8,0,0}
\definecolor{mygreen}{rgb}{0,0.8,0}

\usepackage{pifont}
\newcommand{\good}{{\color{mygreen}\ding{51}}}
\newcommand{\bad}{{\color{myred}\ding{55}}}

\newcommand{\titleabr}{\textit{TriHuman} }

\begin{document}

\title{\titleabr: A Real-time and Controllable Tri-plane Representation for Detailed Human Geometry and Appearance Synthesis}

\author{Heming Zhu}
\affiliation{%
	\institution{Max Planck Institute for Informatics, Saarland Informatics Campus}
	\country{Germany}
}
\email{hezhu@mpi-inf.mpg.de}

\author{Fangneng Zhan}
\affiliation{%
	\institution{Max Planck Institute for Informatics, Saarland Informatics Campus}
	\country{Germany}
}
\email{fzhan@mpi-inf.mpg.de}

\author{Christian Theobalt}
\affiliation{%
	\institution{Max Planck Institute for Informatics, Saarland Informatics Campus and Saarbrücken Research Center for Visual Computing, Interaction and AI}
	\country{Germany}
}
\email{theobalt@mpi-inf.mpg.de}

\author{Marc Habermann}
\affiliation{%
	\institution{Max Planck Institute for Informatics, Saarland Informatics Campus and Saarbrücken Research Center for Visual Computing, Interaction and AI}
	\country{Germany}
}
\email{mhaberma@mpi-inf.mpg.de}

\begin{abstract}
Creating controllable, photorealistic, and geometrically detailed digital doubles of real humans solely from video data is a key challenge in Computer Graphics and Vision, especially when real-time performance is required. 
Recent methods attach a neural radiance field (NeRF) to an articulated structure, e.g., a body model or a skeleton, to map points into a pose canonical space while conditioning the NeRF on the skeletal pose. 
These approaches typically parameterize the neural field with a multi-layer perceptron (MLP) leading to a slow runtime. 
To address this drawback, we propose \titleabr a novel human-tailored, deformable, and efficient tri-plane representation, which achieves real-time performance, state-of-the-art pose-controllable geometry synthesis as well as photorealistic rendering quality. 
At the core, we non-rigidly warp global ray samples into our undeformed tri-plane texture space, which effectively addresses the problem of global points being mapped to the same tri-plane locations. 
We then show how such a tri-plane feature representation can be conditioned on the skeletal motion to account for dynamic appearance and geometry changes.
Our results demonstrate a clear step towards higher quality in terms of geometry and appearance modeling of humans as well as runtime performance.
\end{abstract}


\begin{teaserfigure}
  \includegraphics[width=1.0\textwidth]{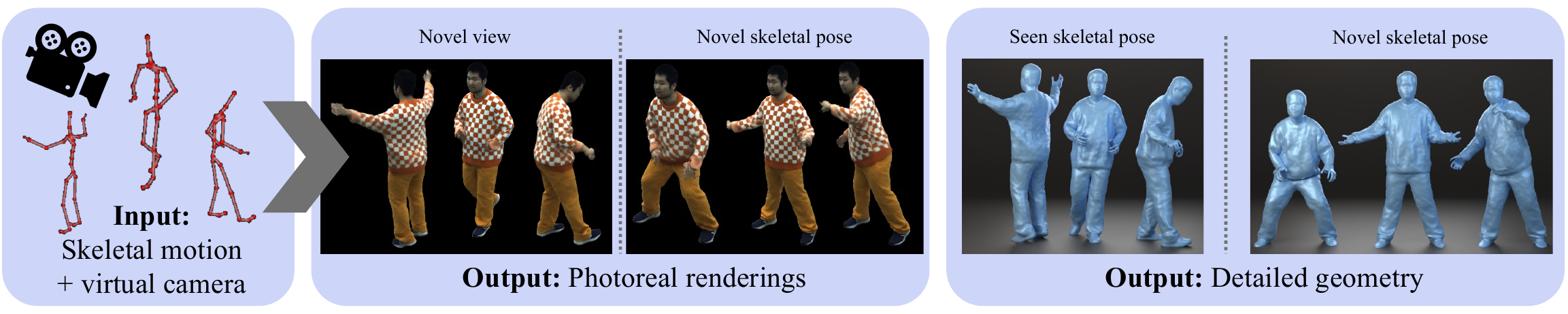}
  \caption{
    \titleabr renders photorealistic images of the virtual human and also generates high-fidelity and topology-consistent clothed human geometry given the skeletal motion and virtual camera view as input.
    Importantly, our method runs in real-time due to our efficient human representation and can be solely supervised on multi-view imagery during training.
  }
  \label{fig:teaser}
\end{teaserfigure}

\maketitle

\blfootnote{Project page: \href{https://vcai.mpi-inf.mpg.de/projects/trihuman}{\color{magenta}{\url{https://vcai.mpi-inf.mpg.de/projects/trihuman}}}}


%
%
\section{Introduction} 
\label{sec:introduction}
\par
Digitizing real humans and creating their virtual double is a long-standing and challenging problem in Graphics and Vision with many applications in the movie industry, gaming, telecommunication, and VR/AR. 
Ideally, the virtual double should be controllable, it should contain highly-detailed and dynamic geometry, and respective renderings should look photoreal while computations should be real-time capable. 
However, so far, creating high-quality and photoreal digital characters requires a tremendous amount of work from experienced artists, takes a lot of time, and is extremely expensive.
Thus, simplifying character creation by learning it directly from multi-view video and making it more efficient has become an active research area in recent years, especially with the advent of deep scene representations.
\par
Recent works~\cite{liu2021neural,ARAH:ECCV:2022,li2022tava} incorporate neural radiance fields (NeRFs) into the modeling of humans due to their capability of representing rich appearance details.
These methods typically map points from a global space, or posed space, into a canonical space by transforming 3D points using the piece-wise rigid transform of nearby bones or surface points of a naked human body model or skeleton.
The canonical point as well as some type of pose conditioning are fed into an MLP parameterizing the NeRF in order to obtain the per-point density and color, which is then volume rendered to obtain the final pixel color.
However, most methods have to perform multiple MLP evaluations per ray, which makes real-time performance impossible. 
\par
To overcome this, we present \titleabr, which is the first real-time method for controllable character synthesis that jointly models detailed, coherent, and motion-dependent surface deformations of arbitrary types of clothing as well as photorealistic motion- and view-dependent appearance.
Given a skeleton motion and camera configuration as input, our method regresses a detailed motion-dependent geometry as well as view- and motion-dependent appearance while for training it only requires multi-view video.
\par 
At the technical core, we represent human geometry and appearance as a signed distance field (SDF) and color field in global space, which can be volume rendered into an image.
To overcome the limited runtime performance of previous methods, we investigate in this work how the efficient tri-plane representation~\cite{Chan2022} can be leveraged to improve runtime performance while maintaining high quality.
Important to note is that the tri-plane representation typically works well for convex shapes like faces as there are only a few points in global space mapping to the same point on the tri-planes. 
However, humans with their clothing are articulated and deformable, which makes it more challenging to prevent \textit{tri-plane mapping collisions}, i.e. global points map to the same tri-plane locations. 
To overcome this, we map global points into an undeformed tri-plane texture space (UTTS) using a deformable human model~\cite{habermann2021real}.
Intuitively, one of the tri-planes coincides with the 2D uv map of the deformable model while the other two planes are perpendicular to the first one and to each other. 
We show that this reduces the mapping collisions when projecting points onto the planes and, thus, leads to better results.
Another challenge is to condition the tri-plane features on the skeletal motion in order to obtain an animatable representation.
Here, we propose an efficient 2D motion texture conditioning encoding the surface dynamics of the deformable model in conjunction with a 3D-aware convolutional architecture~\cite{wang2022rodin} in order to generate tri-plane features that effectively encode the skeletal motion.
Last, these features are decoded to an SDF value and a color using a shallow MLP, and unbiased volume rendering~\cite{wang2021neus} is performed to generate the final pixel color.
\par 
To evaluate our method, we found that most existing datasets contain limited skeletal pose-variation, camera views, and lack ground truth 3D data in order to evaluate the accuracy of the recovered human geometry.
To address this, we propose a new dataset and extend existing datasets consisting of dense multi-view captures using 120 cameras of human performances comprising significantly higher pose variations than current benchmarks.
The dataset further provides skeletal pose annotations, foreground segmentations, and most importantly, 4D ground truth reconstructions.
We demonstrate state-of-the-art results on this novel and significantly more challenging benchmark compared to previous works (see Fig.~\ref{fig:teaser}).
In summary, our contributions are:
\begin{itemize}
    \item{A novel controllable human avatar representation enabling highly detailed and skeletal motion-dependent geometry and appearance synthesis at real time frame rates while supporting arbitrary types of apparel.}
    \item{A mapping, which transforms global points into an undeformed tri-plane texture space (UTTS) greatly reducing the tri-plane collisions.}
    \item{A skeletal motion-dependent tri-plane network architecture encoding the surface dynamics, which allows the tri-plane representation to be skeletal motion conditioned.}
    \item{A new benchmark dataset of dense multi-view videos of multiple people performing various challenging motions, which improves over existing datasets in terms of scale and annotation quality.}
\end{itemize}
\section{Related Works} \label{sec:related}

Recently, neural scene representations \cite{sitzmann2019scene, mildenhall2020nerf, oechsle2021unisurf, wang2021neus, yariv2021volume, yariv2020multiview, DVR} have achieved great success in multifarious vision and graphics applications, including novel view synthesis \cite{yu2021plenoctrees,hedman2021baking,yu2021plenoxels,muller2022instant,chen2022tensorf}, generative modeling \cite{schwarz2020graf,niemeyer2021giraffe,chan2021pi}, surface reconstruction \cite{wang2021neus,oechsle2021unisurf,yariv2021volume}, and many more.
While above works mainly focus on static scenes, recent efforts \cite{tretschk2020nonrigid, park2020nerfies,pumarola2020dnerf,deng2020nasa} has been devoted to extending neural scene / implicit representations for modeling dynamic scenes or articulated objects.
With a special focus on dynamic human modeling, 
existing works can be categorized according to their space canonicalization strategy, which will be introduced in the ensuing paragraphs.

\paragraph{Piece-wise Rigid Mapping.}
Reconstructing the 3D human has attracted increasing attention in recent years.
A popular line of research \cite{alldieck2018detailed, alldieck2018video, xiang2020monoclothcap} utilizes a parametric body model such as SMPL~\cite{loper15} to represent a human body with clothing deformations, which produces an animatable 3D model.
With the emergence of neural scene representations \cite{sitzmann2019scene, mildenhall2020nerf},
series of works \cite{gafni2021dynamic, peng2020neural, su2021anerf, saito2021scanimate,bhatnagar2020loopreg} combine scene representation networks with parametric models~\cite{loper15,blanz1999morphable} to reconstruct dynamic humans.
With a special focus on human body modeling, a number of methods \cite{weng2020vid2actor, chen2021animatable, 2021narf, ARAH:ECCV:2022, bergman2022generative} transform points from a global space, or posed space, into a canonical space by mapping 3D points using piece-wise rigid transformations.
For instance, \citet{chen2021animatable} extend neural radiance fields to dynamic scenes by introducing explicit pose-guided deformation with SMPL to achieve a mapping from the observation space to a constant canonical space;
instead of learning rigid transformations from the full parametric model, NARF~\cite{2021narf} considers only the rigid transformation of the most relevant object part for each 3D point;
ENARF-GAN~\cite{noguchi2022unsupervised} further extend NARF to achieve efficient and unsupervised training from unposed image collections;
to accelerate the neural volume rendering, InstantAvatar~\cite{jiang2022instantavatar} incorporates Instant-NGP~\cite{muller2022instant} to learn a canonical shape and appearance, which derives a continuous deformation field via an efficient articulation module \cite{chen2023fast};
as the inferred geometry from NeRF often lacks detail, ARAH \cite{ARAH:ECCV:2022} builds an articulated signed-distance-field (SDF) representation to better model the geometry of clothed humans, where an efficient joint root-finding algorithm is introduced for the mapping from observation space to canonical space.
However, piece-wise rigid mapping has limited capability to represent complex geometry such as loose clothing.

\begin{figure*}
\centering
  \includegraphics[width=1.\textwidth]{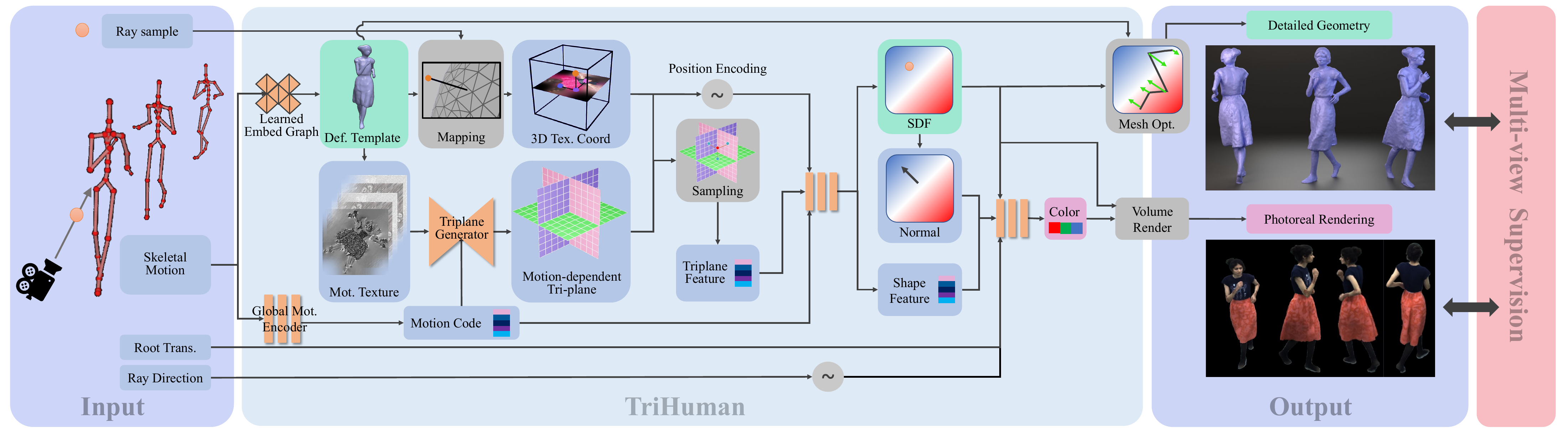}
  \caption{
  \textbf{Overview.}
  Given a skeletal motion and virtual camera view as input, our method generates highly realistic renderings of the human under the specified pose and view.
  To this end, first a rough motion-dependent and deforming human mesh is regressed. 
  From the deformed mesh, we extract several motion features in texture space, which are then passed through a 3D-aware convolutional architecture to generate a motion-conditioned feature tri-plane.
  Ray samples in global space can be mapped into a 3D texture cube, which can be then used to sample a feature from the tri-plane.
  This feature is then passed to a small MLP predicting color and density.
  Finally, volume rendering and our proposed mesh optimization can generate the geometry and images.
  Our method is solely supervised on multi-view imagery.
  }
  \label{fig:pipeline}
\end{figure*}

\paragraph{Piece-wise Rigid and Learned Residual Deformation.}
Recently, an improved deformable NeRF representation \cite{liu2021neural, peng2021animatable, xu2021hnerf, zhang2021stnerf,NNA} has become a common paradigm for dynamic human modeling, by unwarping different poses to a shared canonical space with piece-wise rigid transformations and learned residual deformations~\cite{tretschk2020nonrigid, park2020nerfies, pumarola2020dnerf, zhan2023general, wang2022neural}.
For instance, \citet{liu2021neural} employ an inverse skinning transformation \cite{lewis00,peng2021animatable} to deform the posed space to the canonical pose space, accompanied with a predicted residual deformation for each pose;
similarly, \citet{weng2022humannerf,peng2022animatable} propose to optimize a human representation in a canonical T-pose, relying on a motion field consisting of skeletal rigid and non-rigid deformations;
\citet{wang2022neural} further propose to model the residual deformation by leveraging geometry features and relative displacement;
recently, \citet{li2022tava} incorporate a deformation model that captures non-linear pose-dependent deformations, which is anchored in a LBS formulation;
\citet{instant_nvr} apply the multi-resolution hash encoding to the transformed point and regresses a residual to obtain the canonical space.
While such a residual deformation can typically compensate for smaller misalignments and wrinkle deformations, we found that it typically fails to handle clothing types and deformations that significantly deviate from the underlying articulated structure.

\paragraph{Modeling Surface Deformation.}

Notably, some recent efforts \cite{habermann2021real} have been devoted to modeling both coarse and fine dynamic deformation by introducing a parametric human representation with explicit space-time coherent mesh geometry and high-quality dynamic textures.
However, they still face challenges in capturing fine-scale details due to the complexity of the optimization process involved in deforming meshes with sparse supervision.
Alternatively, the prevailing implicit representation methods offer a more flexible human representation.
\citet{habermann2022hdhumans} propose to condition NeRF on a densely deforming template to enable the tracking of loose clothing and further refine the template deformations.
However, their method requires multiple MLP evaluations per ray sample, resulting in slower computation. Additionally, the recovered surface quality is compromised since they model the scene as a density field rather than a Signed Distance Function (SDF).
Recently, \citet{kwon2023deliffas} achieves real-time rendering of dynamic characters through a surface light field attached to the deformable template mesh. 
Nevertheless, similar to prior methods, the generated geometry is of lower quality and lacks delicate surface details.
%
%
\section{Methodology} 
\label{sec:overview}
Our goal is to obtain a drivable, photorealistic, and geometrically detailed avatar of a real human in any type of clothing solely learned from multi-view RGB video.
More precisely, given a skeleton motion and virtual camera view as input, we want to synthesize highly realistic renderings of the human in motion as well as the high-fidelity and deforming geometry in real time.
An overview of our method is shown in Fig.~\ref{fig:pipeline}.
Next, we define the problem setting (Sec.~\ref{subsec:problem_setting}).
Then, we describe the main challenges of space canonicalization that current methods are facing followed by our proposed space mapping, which alleviates the inherent ambiguities (Sec.~\ref{subsec:mapping}).
Given this novel space canonicalization strategy, we show how this undeformed tri-plane texture space can be efficiently parameterized with a tri-plane representation leading to real-time performance during rendering and geometry recovery (Sec.~\ref{subsec:tri_plane}).
Last, we introduce our supervision and training strategy (Sec.~\ref{subsec:supervision}).
%
%
%
\subsection{Problem Setting} \label{subsec:problem_setting}
\textbf{Input Assumptions.}
We assume a segmented multi-view video of a human actor using $C$ calibrated and synchronized RGB cameras as well as a static 3D template is given. 
$\mathbf{I}_{f,c} \in \mathbb{R}^{H \times W}$ denotes frame $f$ of camera $c$ where $W$ and $H$ are the width and height of the image, respectively.
We then extract the skeletal pose $\boldsymbol{\theta}_f \in \mathbb{R}^P$ for each frame $f$ using markerless motion capture~\cite{captury}.
Here, $P$ denotes the number of degrees of freedom (DoFs).
A skeletal motion from frame $f-k$ to $f$ is denoted as $\boldsymbol{\theta}_{\bar{f}} \in \mathbb{R}^{kP}$ and $\hat{\boldsymbol{\theta}}_{\bar{f}}$ is the translation normalized equivalent, i.e., by displacing the root motion such that the translation of frame $f$ is zero. 
During training, our model takes the skeletal motion as input and the multi-view videos as supervision, while at inference our method only requires a skeletal motion.

%
%
\par
\textbf{Static Scene Representation}.
Recent progress in neural scene representation learning has shown great success in terms of geometry reconstruction~\cite{wang2021neus, neus2} and view synthesis~\cite{mildenhall2020nerf} of \textit{static scenes} by employing neural fields.
Inspired by NeuS~\cite{wang2021neus}, we also represent the human geometry and appearance as neural field $\mathcal{F}_\mathrm{sdf}$ and $\mathcal{F}_\mathrm{col}$:
\begin{equation} \label{eq:static_neural_field_sdf}
    \mathcal{F}_\mathrm{sdf}(p(\mathbf{x}_i); \Gamma)= s_i, \mathbf{q}_i 
\end{equation}
\begin{equation} \label{eq:static_neural_field_col}
    \mathcal{F}_\mathrm{col}(\mathbf{q}_i, s_i, \mathbf{n}_i, p(\mathbf{d}); \Psi)= \mathbf{c}_i 
\end{equation}
where $\mathbf{x}_i \in \mathbb{R}^3$ is a point along the camera $r(t_i,\mathbf{o}, \mathbf{d}) = \mathbf{o} + t_i \mathbf{d}$ with origin $\mathbf{o}$ and direction $\mathbf{d}$. 
$p(\cdot)$ is a positional encoding~\cite{mildenhall2020nerf} to better model and synthesize higher frequency details.
The SDF field stores the SDF $s_i$ and a respective shape code $\mathbf{q}_i$ for every point $\mathbf{x}_i$ in global space.
Note that the normal at point $\mathbf{x}_i$ can be computed as $\mathbf{n}_i =\frac{\partial s_i}{\partial \mathbf{x}_i}$.
Moreover, the color field encodes the color $\mathbf{c}_i$ and as it is conditioned on the viewing direction $\mathbf{d}$, it can also encode view-dependent appearance changes.
In practice, both fields are parameterized as multi-layer perceptrons (MLPs) with learnable weights $\Gamma$ and $\Psi$.
\par 
To render the color of a ray (pixel), volume rendering is performed, which accumulates the color $\mathbf{c}_i$ and the density $\alpha_i$ along the ray as
\begin{equation} \label{eq:volume_render}
\mathbf{c} = \sum^R_i T_i \alpha_i \mathbf{c}_i, \quad T_i = \prod^{i-1}_{j=1} (1 - \alpha_j).
\end{equation}
Here, the density $\alpha_i$ is a function of the SDF.
For an unbiased SDF estimate, the conversion from SDF to density can be defined as 
\begin{equation} \label{eq:alpha}
    \alpha_i = \mathrm{max} \left( \frac{\Phi(s_i) - \Phi(s_{i+1})}{\Phi(s_i)}, 0 \right)
\end{equation}
\begin{equation} \label{eq:psi}
    \Phi(s_i) = (1 + e^{-zx})^{-1},
\end{equation}
where $z$ is a trainable parameter whose reciprocal approaches 0 when training converges.
For a detailed derivation, we refer to the original work~\cite{wang2021neus}.
The scene geometry and appearance can then be solely supervised by comparing the obtained pixel color with the ground truth color, typically employed by an L1 loss.
Important for us, this representation allows modeling of fine geometric details as well as appearance while only requiring multi-view imagery. 
However, for now this representation only allows modeling of static scenes and requires multiple hours of training (even for a single frame).
%
%
\par
\textbf{Problem Setting.}
Instead, we want to learn a dynamic, controllable, and efficient human representation $\mathcal{H}_\mathrm{sdf}$ and 
 $\mathcal{H}_\mathrm{col}$:
\begin{equation} \label{eq:human_neural_field_sdf}
    \mathcal{H}_\mathrm{sdf}(\boldsymbol{\theta}_{\bar{f}}, p(\mathbf{x}_i); \Gamma)= s_{i,f}, \mathbf{q}_{i,f} 
\end{equation}
\begin{equation} \label{eq:human_neural_field_col}
    \mathcal{H}_\mathrm{col}(\boldsymbol{\theta}_{\bar{f}}, \mathbf{q}_{i,f}, s_{i,f}, \mathbf{n}_{i,f}, p(\mathbf{d}); \Psi)= \mathbf{c}_{i,f},
\end{equation}
which is conditioned on the skeletal motion of the human as well. 
Note that the SDF, shape feature, and color are now a function of skeletal motion indicated by the subscript $(\cdot)_f$.
Previous work~\cite{liu2021neural} has shown that naively adding the motion as a function input to the field leads to blurred and unrealistic results.
Many works~\cite{liu2021neural,peng2020neural,peng2021animatable} have therefore tried to transform points into canonical 3D pose space to then query the neural field in this canonical space.
This has shown to improve quality, however, they typically parameterize the field in this space with an MLP leading to slow runtimes. 
\par 
Tri-planes~\cite{Chan2022} offer an efficient alternative and have been applied to generative tasks, however, mostly for convex surfaces such as faces where the mapping onto the planes introduces little ambiguity. 
However, using them for representing the complex, articulated, and dynamic structure of humans in clothing requires additional attention since, if not handled carefully, the mapping onto the tri-plane can lead to so-called mapping collisions, where multiple points in the global space map onto the same tri-plane locations. 
Thus, in the remainder of this section, we first introduce our undeformed tri-plane texture space (UTTS), which effectively reduces these collisions (Sec.~\ref{subsec:mapping}).
Then, we explain how the tri-plane can be conditioned on the skeletal motion using an efficient encoding of surface dynamics into texture space, which is then decoded into the tri-plane features leveraging a 3D-aware convolutional architecture~\cite{wang2022rodin} (Sec.~\ref{subsec:tri_plane}).
Last, we describe our supervision and training strategy (Sec.~\ref{subsec:supervision}).

%
%
\subsection{Undeformed Tri-plane Texture Space} \label{subsec:mapping}
Intuitively, our idea is that one of the tri-planes, i.e., the \textit{surface plane}, corresponds to the surface of a skeletal motion-conditioned deformable human mesh model, while the other two planes, i.e., the \textit{perpendicular planes}, are perpendicular to the first one and to each other.
Next, we define the deformable and skeletal motion-dependent surface model of the human.
%
%
\par
\textbf{Motion-dependent and Deformable Human Model.}
We assume a person-specific, rigged and skinned triangular mesh with $N$ vertices $\mathbf{M} \in \mathbb{R}^{N \times 3}$ is given and the vertex connectivity remains fixed.
The triangular mesh $\mathbf{M}$ is obtained from a 3D scanner ~\cite{treedys2020} and down-sampled to around $5{,}000$ vertices to strike a balance between quality and efficiency.
Now, we denote the deformable and motion-dependent human model as 
\begin{equation} \label{eq:our_model}
 \mathcal{V}(\boldsymbol{\theta}_{\bar{f}};\Omega) = \mathbf{V}_{\bar{f}}
\end{equation}
where $\Omega \in \mathbb{R}^{W}$ are the learnable network weights and $\mathbf{V}_{\bar{f}} \in \mathbb{R}^{N \times 3}$ are the posed and non-rigidly deformed vertex positions.
Important for us, this function has to satisfy two properties:
1) It has to be a function of skeletal motion.
2) It has to have the capability of capturing non-rigid surface deformation.
\par 
We found that the representation of \citet{habermann2021real} meets these requirements, and we, thus, leverage it for our task. 
In their formulation, the human geometry is first non-rigidly deformed in a canonical pose as
\begin{equation} \label{eq:ddc_deform}
\mathbf{Y}_{v} = \mathbf{D}_i + \sum_{k \in \mathcal{N}_{\mathrm{vn},v}} \mathbf{w}_{v,k}(R(\mathbf{A}_k)(\mathbf{M}_v - \mathbf{G}_k) + \mathbf{G}_k + \mathbf{T}_k)
\end{equation}
where $\mathbf{M}_v \in \mathbb{R}^{3}$ and $\mathbf{Y}_v \in \mathbb{R}^{3}$ denote the undeformed and deformed template vertices in the rest pose. $\mathcal{N}_{\mathrm{vn},v} \in \mathbb{N}$ denotes the indexes of embedded graph nodes that are connected to template mesh vertex $v \in \mathbb{R}^{3}$.
$\mathbf{G}_k \in \mathbb{R}^{3}$, $\mathbf{A}_k \in \mathbb{R}^{3}$, and $\mathbf{T}_k \in \mathbb{R}^{3}$ indicate the rest positions, rotation Euler angles, and translations of the embedded graph nodes. 
Specifically, the connectivity of the embedded graph $\mathbf{G}_k$ can be obtained by simplifying the deformable template mesh $\mathbf{M}$ with quadric edge collapse decimation in Meshlab ~\cite{meshlab}.
$R(\cdot)$ denotes the function that converts the Euler angle to a rotation matrix.
Similar to ~\cite{sorkine2007rigid}, we compute the weight applied to the neighboring vertices $\mathbf{w}_{v,k} \in \mathbb{R}$ based on geodesic distances.

To model higher-frequency deformations, an additional per-vertex displacement $\mathbf{D}_i \in \mathbb{R}^{3}$ is added.
The embedded graph parameters $\mathbf{A},\mathbf{T}$, and per-vertex displacements $\mathbf{D}$ are further functions of translation-normalized skeletal motion implemented as two graph convolutional networks $\mathcal{F_\mathrm{eg}}$ and $\mathcal{F_\mathrm{delta}}$:
\begin{align}\label{eq:gcn}
\mathcal{F_\mathrm{eg}}(\hat{\boldsymbol{\theta}}_{\bar{f}}; \Omega_\mathrm{eg}) &= \mathbf{A}, \mathbf{T} \\
\mathcal{F_\mathrm{delta}}(\hat{\boldsymbol{\theta}}_{\bar{f}}; \Omega_\mathrm{delta}) &= \mathbf{D}
\end{align}
where the skeletal motion is encoded according to \cite{habermann2021real}.
For more details, we refer to the original work.

Finally, the deformed vertices $\mathbf{Y}_{v}$ in the rest pose can be posed using Dual Quaternion (DQ) skinning $\mathcal{S}$~\cite{kavan2007skinning}, which defines the motion-dependent deformable model
\begin{equation} \label{eq:dq}
\mathcal{S}(\boldsymbol{\theta},\mathbf{Y}) = \mathbf{V}_{\bar{f}} = \mathcal{V}(\boldsymbol{\theta}_{\bar{f}};\Omega).
\end{equation}
Note that Eq.~\ref{eq:dq} is 1) solely a function of the skeletal motion and 2) can account for non-rigid deformations by means of training the weights $\Omega$ and, thus, this formulation satisfies our initial requirements.
\par
\begin{figure}[t]
\includegraphics[width=1.0\linewidth]{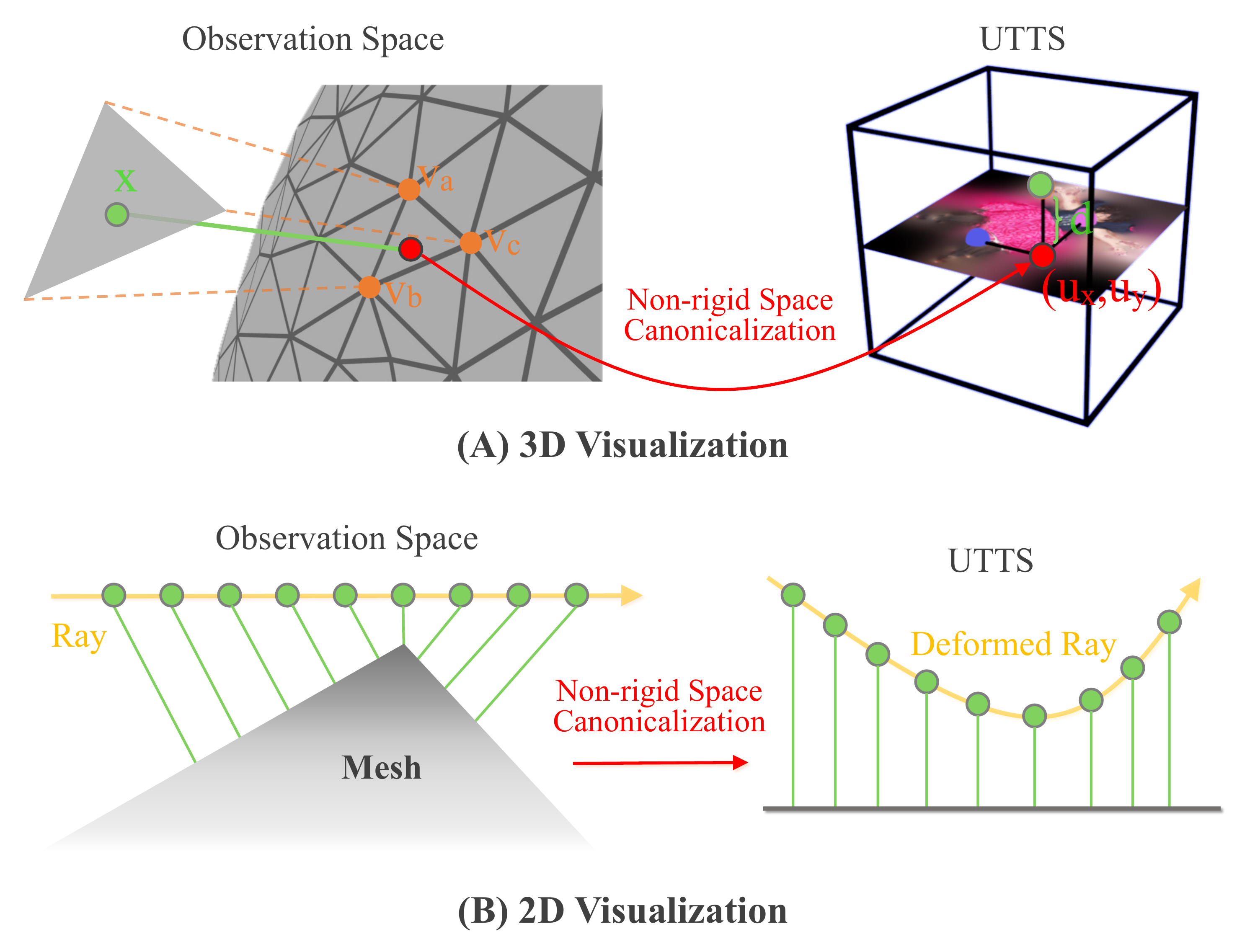}
  \caption{
        The illustration of UTTS mapping from 3D perspective (A) and 2D perspective (B). Each spatial sample in the observation space undergoes a non-rigid transformation into the UTTS space via non-rigid canonicalization.
  }
  \label{fig:supplutts}
\end{figure}

\textbf{Non-rigid Space Canonicalization.}
Next, we introduce our non-rigid space canonicalization function (see also Fig.~\ref{fig:supplutts})
\begin{equation} \label{eq:map}
    \mathcal{M}(\mathcal{V}(\boldsymbol{\theta}_{\bar{f}};\Omega), \mathbf{x}) = \bar{\mathbf{x}},
\end{equation}
which takes the deformable template and a point $\mathbf{x}$ in global space and maps it to the so-called undeformed tri-plane texture space, denoted as $\bar{\mathbf{x}}$, as explained in the following.
Given the point $\mathbf{x}$ in global space and assuming the closest point $\mathbf{p}$ is located on the non-degenerated triangle with vertices $\{\mathbf{v}_a, \mathbf{v}_b, \mathbf{v}_c \}$ on the posed and deformed template $\mathbf{V}_{\bar{f}}$, the closest point on the surface can either be a face, edge, or vertex on the mesh. 
In the following, we discuss the different cases where the goal is to find the UV coordinate of $\mathbf{p}$ as well as the distance between $\mathbf{x}$ and $\mathbf{p}$, which then defines the 3D coordinate $\bar{\mathbf{x}}$ in UTTS.
\par 
\textit{1) Face.}
If the closest point lies on the triangular surface, the distance $d$ and 2D texture coordinate $\mathbf{u}$ can be computed as 
\begin{equation}
\begin{aligned}
 \mathbf{p} &= \mathbf{x} - (\mathbf{n}_f \cdot (\mathbf{x} - \mathbf{v}_a)) \mathbf{n}_f\\ 
 \lambda_a &= \frac{\lVert (\mathbf{v}_c - \mathbf{v}_b) \times (\mathbf{p} - \mathbf{v}_b) \rVert _{2}}{\lVert (\mathbf{v}_c - \mathbf{v}_b) \times (\mathbf{v}_a - \mathbf{v}_b) \rVert _{2}}\\
 \lambda_b &= \frac{\lVert (\mathbf{v}_a - \mathbf{v}_c) \times (\mathbf{p} - \mathbf{v}_c) \rVert _{2}}{\lVert (\mathbf{v}_c - \mathbf{v}_b) \times (\mathbf{v}_a - \mathbf{v}_b) \rVert _{2}}\\
  d &= \lVert \mathbf{x} - \mathbf{p} \rVert _{2}\\
 \mathbf{u} &= \lambda_a \mathbf{u}_a + \lambda_b \mathbf{u}_b + (1 - \lambda_a - \lambda_b) \mathbf{u}_c
\end{aligned}
\end{equation}
where $\mathbf{n}_f$ denotes the face normal of the closest surface, and $\mathbf{u}_a$, $\mathbf{u}_b$ and $\mathbf{u}_c$ indicates the UV coordinates of the triangle vertices.
\par 
\textit{2) Edge.}
For global points mapping onto the edge $(\mathbf{v}_a, \mathbf{v}_b) $, the uv coordinate and distance can be computed as:
\begin{equation}
\begin{aligned}
 \lambda &= \frac{(\mathbf{v}_b - \mathbf{v}_a) \cdot (\mathbf{x} - \mathbf{v}_a)}{\lVert (\mathbf{v}_b - \mathbf{v}_a) \rVert _{2}}\\
 \mathbf{p} &= \mathbf{v}_a + \lambda (\mathbf{v}_b - \mathbf{v}_a) \\
 d &= \lVert \mathbf{x} - \mathbf{p} \rVert _{2} \\
 \mathbf{u} &= \lambda\mathbf{u}_a + (1 - \lambda)\mathbf{u}_b.
\end{aligned}
\end{equation}
\par 
\textit{3) Vertex.}
If the global point $\mathbf{x}$ maps onto a vertex $\mathbf{v}_a$, the uv coordinate and the distance to the template can be computed with:
\begin{equation}
\begin{aligned}
 d &= \lVert \mathbf{x} - \mathbf{v}_a \rVert _{2} \\
 \mathbf{u} &= \mathbf{u}_a.
\end{aligned}
\end{equation}
\par 
So far, we can now canonicalize points from global 3D to our UTTS space and we denote the canonical point $(\mathbf{u},d)^T$ or $(u_x,u_y,d)^T$ simply as $\bar{\mathbf{x}}$.
Note that $\mathbf{u}=(u_x, u_y)$ denotes the point on our surface plane and $(u_x,d)$ and $(u_y,d)$ correspond to the points on the perpendicular planes.
These coordinates can now be used to query the features on the respective tri-planes.
\par 
Concerning the mapping collision, we highlight that case 1) where a point maps onto a triangle is a bijection and, thus, the concatenated tri-plane features are unique, which was our goal.
Only for case 2) and 3) the aforementioned collisions can happen since the uv coordinate on the surface is no longer unique for points with the same distance to a point on the edge or the mesh vertex itself.
However, the occurrence of these cases highly depends on how far away from the deformable surface points are still being sampled.
By constraining the maximum distance to $d_\mathrm{max}$, which effectively means we only draw samples close to the deformable surface, we found that case 2) and 3) happens less frequently. 
However, when the deformable model is not aligned, this will introduce an error by design as surface points are not sampled in regions covered by the human.
Therefore, we gradually deform the surface along the SDF field to account for such cases, and iteratively reduce $d_\mathrm{max}$.
In the limit, this strategy reduces the mapping collisions, improves sampling efficiency, and ensures that the sampled points do not miss the real surface.
More details about this will be explained in Sec.~\ref{subsec:supervision} and in our supplemental material.

%
%
\subsection{Efficient and Motion-dependent Tri-plane Encoding} 
\label{subsec:tri_plane}
So far, we are able to map points in the global space to our UTTS space, however, as mentioned earlier we want to ensure that the tri-planes contain skeletal motion-aware features.
Thus, we propose a 3D-aware convolutional motion encoder:
\begin{equation} \label{eq:motion_encoder}
\mathcal{E}(\mathbf{T}_{\mathrm{p},f}, \mathbf{T}_{\mathrm{v},f}, \mathbf{T}_{\mathrm{a},f}, \mathbf{T}_{\mathrm{u},f}, \mathbf{T}_{\mathrm{n},f}, \mathbf{g}_f; \Phi ) = \mathbf{P}_{x,f},\mathbf{P}_{y,f},\mathbf{P}_{z,f},
\end{equation}
which takes several 2D textures as input, encoding the position $\mathbf{T}_{\mathrm{p},f}$, velocity $\mathbf{T}_{\mathrm{v},f}$, acceleration $\mathbf{T}_{\mathrm{a},f}$, uv coordinate $\mathbf{T}_{\mathrm{u},f}$, and normal $\mathbf{T}_{\mathrm{n},f}$ of the deforming human mesh surface, 
which we root normalize, i.e., we subtract the skeletal root translation from the mesh vertex positions (Eq.~\ref{eq:dq}) and scale them to a range of $[-1, 1]$.
Note that the individual texel values for $\mathbf{T}_{\mathrm{v},f}$, $\mathbf{T}_{\mathrm{a},f}$, $\mathbf{T}_{\mathrm{u},f}$ and $\mathbf{T}_{\mathrm{n},f}$ can be simply computed using inverse texture mapping.
The first 3 textures encode the dynamics of the deforming surface, the UV map encodes a unique ID for each texel covered by a triangle in the uv-atlas, and the normal texture emphasizes the surface orientation. 
All textures have a resolution of $256 \times 256$. 
Here, $\mathbf{g}_f$ is a global motion code, which is obtained by encoding the translation normalized motion vector $\hat{\boldsymbol{\theta}}_{\bar{f}}$ through a shallow MLP. 
Notably, the global motion code provides awareness of global skeletal motion and, thus, is able to encode global shape and appearance effects, which may be hard to encode through the above texture inputs.
\par 
Given the motion texture and the global motion code, we first adopt three separated convolutional layers to generate the coarse level / initial features for each plane in the tri-plane. 
Inspired by the design of \citet{wang2022rodin}, we adopt a 5-layer UNet with roll-out convolutions to fuse the features from different planes, which enhances the spatial consistency in the feature space. 
Moreover, we concatenate the global motion code channel-wise to the bottleneck feature maps to provide an awareness of the global skeletal motion. Please refer to the supplemental materials for more details regarding the network architectures of the 3D-aware convolutional motion encoder and the global motion encoder.
\par 
Finally, our motion encoder outputs three orthogonal skeletal motion-dependent tri-planes $\mathbf{P}_{x,f}$, $\mathbf{P}_{y,f}$, and $\mathbf{P}_{z,f}$.
The tri-plane feature for a sample $\bar{\mathbf{x}}_i$ in UTTS space can be obtained by querying the planes $\mathbf{P}_{x,f},\mathbf{P}_{y,f},\mathbf{P}_{z,f}$ at $\mathbf{u} = (u_x, u_y)$, $(u_x,d)$, and $(u_y,d)$, respectively, thanks to our UTTS mapping.
The final motion-dependent tri-plane feature $\mathbf{F}_{i,f}$ for a sample $\bar{\mathbf{x}}_i$ can then be obtained by concatenating the three individual features from each plane.
Finally, our initial human representation in Eq.~\ref{eq:human_neural_field_sdf} and \ref{eq:human_neural_field_col}
can be re-defined with the proposed efficient motion-dependent triplane as
\begin{equation} \label{eq:our_final_sdf}
\mathcal{H}_\mathrm{sdf}(\mathbf{F}_{i,f}, \mathbf{g}_f, p(\bar{\mathbf{x}}_i); \Gamma) = s_{i,f}, \mathbf{q}_{i,f}
\end{equation}
\begin{equation} \label{eq:our_final_col}
\mathcal{H}_\mathrm{col}(\mathbf{q}_{i,f}, s_{i,f}, \mathbf{n}_{i,f}, p(\mathbf{d}), \mathbf{t}_f; \Psi) = \mathbf{c}_{i,f}.
\end{equation}
Here, $\mathbf{t}_f$ is the global position of the character accounting for the fact that appearance can change depending on the global position in space due to non-uniform lighting conditions.
In practice, the above functions are parameterized by two shallow (4-layer) MLPs with a width of 256, since most of the capacity is in the tri-plane motion encoder whose evaluation time is \textit{independent} of the number of samples along a ray.
Thus, evaluating a single sample $i$ can be efficiently performed leading to real-time performance.
%
%
\subsection{Supervision and Training Strategy} \label{subsec:supervision}
First, we pre-train the deformable mesh model (Eq.~\ref{eq:our_model}) according to \cite{habermann2021real}.
Then, the training of our human representation proceeds in 3 stages. 
We refer to the supplemental materials for the implementation details of the loss terms.
\par
\textbf{Field Pre-Training.}
Given the initial deformed mesh, we train the SDF (Eq.~\ref{eq:our_final_sdf}) and color field (Eq.~\ref{eq:our_final_col}) using the following losses:
\begin{equation} \label{eq:stage1}
\mathcal{L}_\mathrm{col} + \mathcal{L}_\mathrm{mask} + \mathcal{L}_\mathrm{eik} + \mathcal{L}_\mathrm{seam}.
\end{equation}
Here, the $\mathcal{L}_\mathrm{col}$ and $\mathcal{L}_\mathrm{mask}$ denote an L1 color and mask loss ensuring that the rendered color for a ray matches the ground truth one, and that accumulated transmittance along the ray coincides with the ground truth masks. 
Moreover, the Eikonal loss $\mathcal{L}_\mathrm{eik}$~\cite{eikonal} regularizes the network predictions for the SDF value.
Last, we introduce a seam loss $\mathcal{L}_\mathrm{seam}$, which samples points along texture seams on the mesh.
For a single point on the seam, the two corresponding uv coordinates in the 3D texture space are computed and both are randomly displaced along the third dimension resulting in two samples, where the loss ensures that the SDF network predicts the same value for both points.
This ensures that the SDF prediction on a texture seam is consistent.
More details about the seam loss are provided in the supplemental document.
\par 
\textbf{SDF-driven Surface Refinement.}
Once the SDF and color field training converged, we further refine the pre-trained deformable mesh model to better align with the SDF using the following loss terms:
\begin{equation} \label{eq:stage2}
\mathcal{L}_\mathrm{sdf} + \mathcal{L}_\mathrm{reg} + \mathcal{L}_\mathrm{zero} + \mathcal{L}_\mathrm{normal} + \mathcal{L}_\mathrm{area}.
\end{equation}
The SDF loss $\mathcal{L}_\mathrm{sdf}$ ensures that the SDF queried at the template vertex positions is equal to zero, thus, dragging the mesh towards the implicit surface estimate of the network.
Though this term could also backpropagate into the mapping directly, i.e. into the morphable clothed human body model, we found network training is more stable when keeping the mapping fixed according to the initial deformed mesh.
$\mathcal{L}_\mathrm{reg}$ denotes the Laplacian loss that penalizes the Laplacian difference between the updated posed template vertices and the posed template vertices before the surface refinement. $\mathcal{L}_\mathrm{zero}$ denotes a smoothing term that pushes the Laplacian for the template vertices to approach zero. As the flipping of the faces would lead to abrupt changes in the UV parameterization, we adopt a face normal consistency loss $\mathcal{L}_\mathrm{normal}$ to avoid the face flipping, which can be computed through the cosine similarity of neighboring face normals. Moreover, as the degraded faces would lead to numerical errors for UV mapping, we adopted the face stretching loss $\mathcal{L}_\mathrm{area}$, which can be computed with the deviation of the edge lengths within each face.
Again, more details about the individual loss terms can be found in the supplemental document.
Importantly, the more SDF-aligned template will allow us to adaptively lower the maximum distance $d_\mathrm{max}$ for the tri-plane dimension orthogonal to the UV layout \textit{without} missing the real surface while also reducing mapping collisions.
\par 
\textbf{Field Finetuning.}
Since the deformable mesh is now updated, the implicit field surrounding it has to be updated accordingly. Therefore, in our last stage, we once more refine the SDF and color field using the following losses
\begin{equation} \label{eq:stage3}
\mathcal{L}_\mathrm{col} + \mathcal{L}_\mathrm{mask} + \mathcal{L}_\mathrm{eik} + \mathcal{L}_\mathrm{seam} + \mathcal{L}_\mathrm{lap} + \mathcal{L}_\mathrm{perc}.
\end{equation}
while lowering the distance $d_\mathrm{max}$, which effectively reduces mapping collisions.
This time we also add a patch-based perceptual loss $\mathcal{L}_\mathrm{lap}$ ~\cite{zhang2018unreasonable} and a laplacian pyramid loss $\mathcal{L}_\mathrm{perc}$ ~\cite{bojanowski2019optimizing}.
We found that this helps to improve the detailedness in terms of appearance and geometry.
%
%
\par 
\textbf{Real-time Mesh Optimization.}
At test time, we propose a real-time mesh optimization, which embosses the template mesh with fine-grained and motion-aware geometry details from the implicit field.
We subdivide the original template once using edge subdivision, i.e., cutting the edges into half, to obtain a higher resolution output. Then, we update the subdivided template mesh along the implicit field, i.e., by evaluating the SDF value and displacing the vertex along the normal by the magnitude of the SDF value. 
Due to our efficient SDF evaluation leveraging our tri-planar representation, this optimization is very efficient, allowing real-time generation of high-quality and consistent clothed human geometry.
%
%
\par 
\textbf{Implementation Details.}
Our approach is implemented in PyTorch~\cite{paszke2019pytorch} and custom CUDA kernels. 
Specifically, we implement the skeletal computation, rasterization-based ray sample filter, and mapping with custom CUDA kernels. The remaining components are implemented using PyTorch. 
We train our method on a single Nvidia A100 graphics card using ground truth images with a resolution of $1285 \times 940$.
The Field Pre-Training stage is trained for 600K iterations using a learning rate of 5e-4 scheduled with a cosine decay scheduler, 
which takes around 2 days. Here, we set the distance $d=4cm$.
We perform a random sampling of $4{,}096$ rays from the foreground pixels of the ground truth images. Along each of these rays, we take 64 samples for ray marching. 
The loss terms adopted for supervising the training of Field Pre-Training (Eq.~\ref{eq:stage1}) stage are weighted as 1.0, 0.1, 0.1, and 1.0 in the order of appearance in the equation. 
The SDF-driven Surface Refinement stage is trained for 200k iterations using a learning rate of 1e-5, which takes around 0.4 days.
Here, the losses (Eq.~\ref{eq:stage2}) are weighted with 1.0, 0.15, 0.005, 0.005, 5.0, again in the order of appearance in the equation.
Last, the Field Finetuning stage is trained for 300k iterations with a learning rate of 2e-4, decayed with a cosine decay scheduler,
which takes around 1.4 days.
Here, we set the distance $d=2cm$. 
Similar to the Field Pre-Training stage, we again randomly sampled 4096 rays from the foreground pixels and adopted 64 samples per ray for ray marching. Moreover, we randomly crop patches with a resolution of $128$ for evaluating perceptual-related losses, i.e., $\mathcal{L}_\mathrm{lap}$, and $\mathcal{L}_\mathrm{perc}$.
This time, the losses (Eq.~\ref{eq:stage3}) are weighted with 1.0, 0.1, 0.1, 1.0, 1.0, 0.5.
\begin{figure*}
\centering
\includegraphics[width=\linewidth]{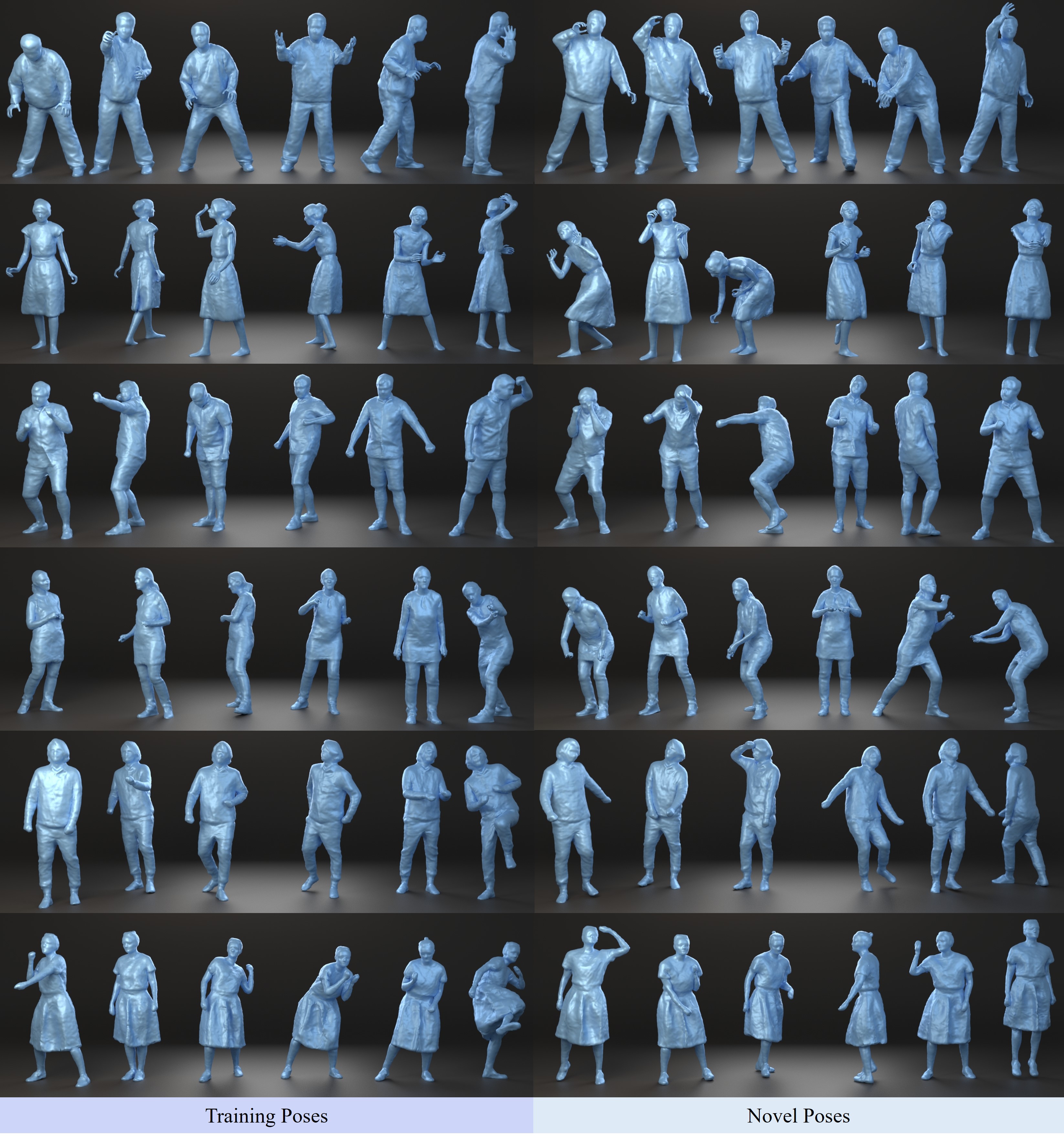}
\caption{
\textbf{Qualitative geometry results.}
We show geometry synthesis results of our method for training and novel skeletal motions. 
Note that in both cases, our method generates high-fidelity geometry in real-time. 
This can be especially observed in the clothing areas where dynamic wrinkles are forming as a function of the skeletal motion.
}
\label{fig:quali_geometry}
\end{figure*}
\begin{figure*}
\centering
\includegraphics[width=\linewidth]{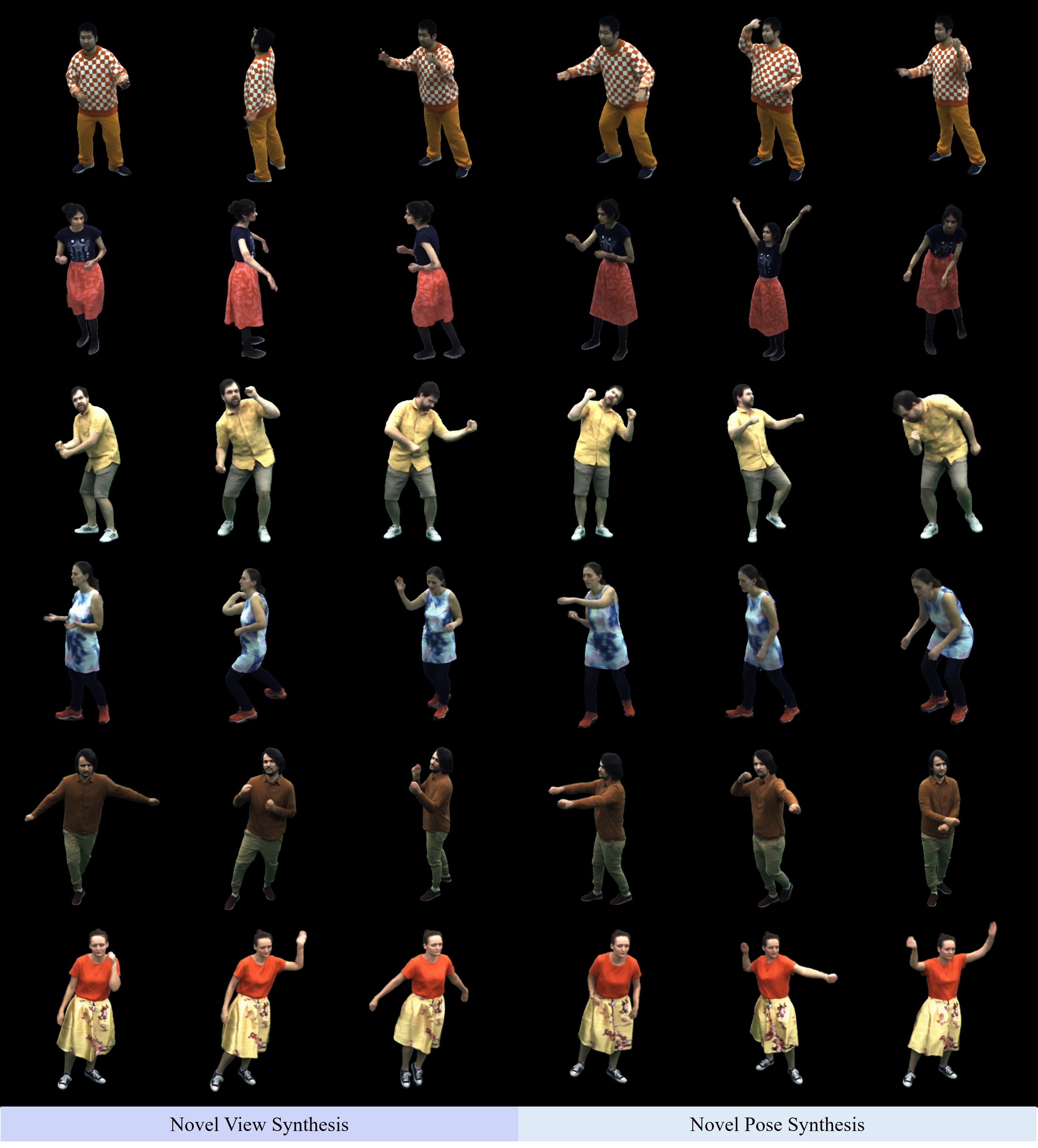}
\caption{
\textbf{Qualitative image synthesis results.}
We show results of our method in terms of image synthesis. 
Our method achieves photorealistic renderings of virtual humans in real time.
We demonstrate high visual quality for, both, novel views and skeletal motions.
Please note how the appearance dynamically changes, given different views and poses.
}
\label{fig:quali_synthesis}
\end{figure*}
%
%
\section{Dataset} \label{sec:dataset}
Our new dataset comprises three subjects wearing different types of apparel. 
We recorded two separate sequences for training and testing for each subject where the person is performing various challenging motions. We assume there is no other person or object in the capture volume during recording. Furthermore, hand-cloth interaction is also avoided throughout the recording process.
The training sequences typically contain $30{,}000$ frames, and the testing sequence around $7{,}000$ frames. 
We record the sequences with a multi-camera system consisting of 120 synchronized and calibrated cameras at a framerate of $25$ fps. 
Notably, to assess the generalization capability of TriHuman, we selected sequences from $4$ cameras for testing and adopted the remaining sequences for training.
For all frames, we provide skeletal pose tracking using markerless motion capture~\cite{captury}, foreground segmentations using background matting~\cite{BMSengupta20}, and ground truth 3D geometry, which we obtained with the state-of-the-art implicit surface reconstruction method~\cite{neus2}.
Note that we use the ground truth geometry solely for evaluating our method.
\par 
Moreover, we reprocessed three subjects from the \textit{DynaCap}~\cite{habermann2021real} dataset, which is publicly available.
More specifically, we improve the foreground segmentations and also provide ground truth geometry for each frame. 
\par 
To the best of our knowledge, there is no other dataset available that has similar specifications, i.e., very long sequences for individual subjects in conjunction with 3D ground truth meshes. 
Thus, we believe this dataset can further stimulate research in this important direction.
\begin{table*}[h]
    \renewcommand\tabcolsep{8.0pt}
\small
    \centering
    \caption{\textbf{Quantitative view synthesis comparison.}
    We quantitatively compare \titleabr to other methods on, both, tight and loose type of apparel on \textit{seen} skeletal motions.
    We highlight the \colorbox{red}{best}, \colorbox{orange}{second-best}, and \colorbox{yellow}{third-best} scores.
    We consistently outperform previous real-time methods in all metrics.
    Concerning offline approaches, we demonstrate superior geometric quality and the highest PSRN metrics.
    In terms of perceptual metrics, we achieve better or slightly worse results. However, our method achieves real-time performance.
    }
    \label{tab:comp_view_rec}
    \begin{tabular}{l|c|ccc|ccc}
    \hline
    & & \multicolumn{3}{|c|}{\textit{Tight Clothing}} & \multicolumn{3}{|c}{\textit{Loose Clothing}} \\
    \cline{3-8}
    \multirow{-2}{*}{\textbf{Methods}} & \multirow{-2}{*}{\textbf{Real-time}} & \textbf{PSNR} $\uparrow$  & \textbf{LPIPS} $\downarrow$ & \textbf{Chamfer} $\downarrow$ & \textbf{PSNR} $\uparrow$  & \textbf{LPIPS} $\downarrow$ & \textbf{Chamfer} $\downarrow$ \\
    \hline
    NA~\cite{liu2021neural}        &   \bad   & 30.33 & 23.71& \cellcolor{yellow} 1.751  & 25.30 & 50.01 & 5.072  \\
    TAVA~\cite{li2022tava}       &  \bad   & 24.61 & 62.26  & 7.814 & 27.31  & 37.55 &  4.717 \\
    HDHumans~\cite{habermann2022hdhumans}   &  \bad   & \cellcolor{yellow} 30.98 & \cellcolor{red} 15.09 & \cellcolor{orange} 1.622 & 
\cellcolor{orange} 29.24 & \cellcolor{red} 15.79  & \cellcolor{orange} 2.596  \\
    DDC~\cite{habermann2021real} & \good & \cellcolor{orange} 31.21  & \cellcolor{yellow} 22.56 & 2.064 & \cellcolor{yellow} 28.10  & \cellcolor{yellow} 31.68 &  \cellcolor{yellow} 2.836 \\
    \textbf{Ours} & \good  & \cellcolor{red} 32.78   & \cellcolor{orange} 18.75  & \cellcolor{red} 1.007  & 
\cellcolor{red} 31.68 & \cellcolor{orange} 16.15 & \cellcolor{red} 1.488 \\
    \hline
    \end{tabular}
\end{table*}

\begin{table*}[h]
    \renewcommand\tabcolsep{8.0pt}
\small
    \centering
    \caption{\textbf{Quantitative pose synthesis comparison.}
    Here, we quantitatively compare our method with prior works for, both, loose and tight types of clothing on \textit{novel} skeletal motions. 
    Note that our real-time method consistently achieves the lowest geometric error compared to all other works, also including \textit{offline} methods.
    In terms of image synthesis, we outperform other real-time methods in all metrics.
    Concerning offline approaches, we report the best PSNR result while having slightly lower LPIPS scores.
    }
    \label{tab:comp_pose_geo}
    \begin{tabular}{l|c|ccc|ccc}
    \hline
    &  & \multicolumn{3}{c|}{\textit{Tight Clothing}} & \multicolumn{3}{|c}{\textit{Loose Clothing}} \\
    \cline{3-8}
    \multirow{-2}{*}{\textbf{Methods}} & \multirow{-2}{*}{\textbf{Real-time}} 
    & \textbf{PSNR} $\uparrow$ & \textbf{LPIPS} $\downarrow$ & \textbf{Chamfer} $\downarrow$ & \textbf{PSNR} $\uparrow$ & \textbf{LPIPS} $\downarrow$ & \textbf{Chamfer} $\downarrow$ \\
    \hline
    NA~\cite{liu2021neural}&  \bad & \cellcolor{orange} 28.78 & \cellcolor{yellow} 25.78 & \cellcolor{orange} 1.986  & 25.03 & 44.20 & 4.792   \\
    TAVA~\cite{li2022tava}   &     \bad    & \cellcolor{yellow} 28.30 & 37.47 & 7.957 &  26.31  & 50.11  & 4.826 \\
    HDHumans~\cite{habermann2022hdhumans}   &  \bad   & 28.17  & \cellcolor{red} 20.69  & \cellcolor{yellow} 2.082 & \cellcolor{orange} 26.71 & \cellcolor{orange} 22.75  & \cellcolor{orange} 3.424  \\
    DDC~\cite{habermann2021real}     &   \good     & 27.77  & 30.16  & 2.428 & \cellcolor{yellow} 26.43 & \cellcolor{yellow} 32.22 & \cellcolor{yellow} 3.532  \\
    \textbf{Ours} & \good  & \cellcolor{red} 29.61  & \cellcolor{orange} 23.73  & \cellcolor{red} 1.686 & \cellcolor{red} 27.58  & \cellcolor{red} 22.65  & \cellcolor{red} 2.743  \\
    \hline
    \end{tabular}
\end{table*}

%
%
\section{Experiments} \label{sec:experiments}
We first provide qualitative results of our approach concerning geometry synthesis and appearance synthesis (Sec.~\ref{subsec:qualitative}). 
Then, we compare our method with prior works focusing on the same task (Sec.~\ref{subsec:comparisons}).
Last, we ablate our major design choices, both, quantitatively and qualitatively (Sec.~\ref{subsec:ablation}).
%
%
%
\subsection{Qualitative Results} 
\label{subsec:qualitative}
For qualitative evaluation of our method, we selected six subjects wearing different types of apparel, ranging from very loose types of apparel such as dresses to more tight clothing such as short pants and T-shirts. 
Three subjects are from our newly acquired dataset and three subjects are from the publicly available DynaCap dataset~\cite{habermann2021real}.
All sequences contain a large range of different poses making it especially challenging compared to previous datasets where pose variation is rather limited~\cite{peng2020neural}.
\par 
\textbf{Geometry Synthesis.}
We qualitatively evaluate the geometry reconstruction performance of our model as shown in Fig.~\ref{fig:quali_geometry}.
For subjects with various types of apparel, our model allows us to reconstruct the high-fidelity geometry faithfully, including the challenging areas with dynamic wrinkles.
Note that the high-fidelity geometry and geometric details are dynamically changing as a function of the skeletal motion. 
This can be best observed in our supplemental video.
Importantly, our model is capable of generating such high-fidelity results in real-time and yields consistent performance for both training poses and novel poses. 
Moreover, our recovered geometry is in correspondence over time, making it well-suited for applications such as consistent texture augmentation (Sec.~\ref{sec:application}).
\par 
\textbf{Image Synthesis.}
Additionally, we show the qualitative results of our method for image synthesis in Fig.~\ref{fig:quali_synthesis} for the same subjects. 
Our model yields highly photorealistic renderings of the entire human in real time for, both, novel views and novel poses, which significantly deviate from the ones seen during training.
Notably, view-dependent appearance effects, small clothing wrinkles, and loose clothing are also synthesized realistically for all clothing types.
Again, we refer to the supplemental video for more results.
\par 
These results demonstrate the versatility and capabilities of our approach in terms of geometry recovery and synthesis as well as photorealistic appearance modeling enabling novel view synthesis as well as novel pose synthesis.
\begin{figure*}
\centering
\includegraphics[width=\linewidth]{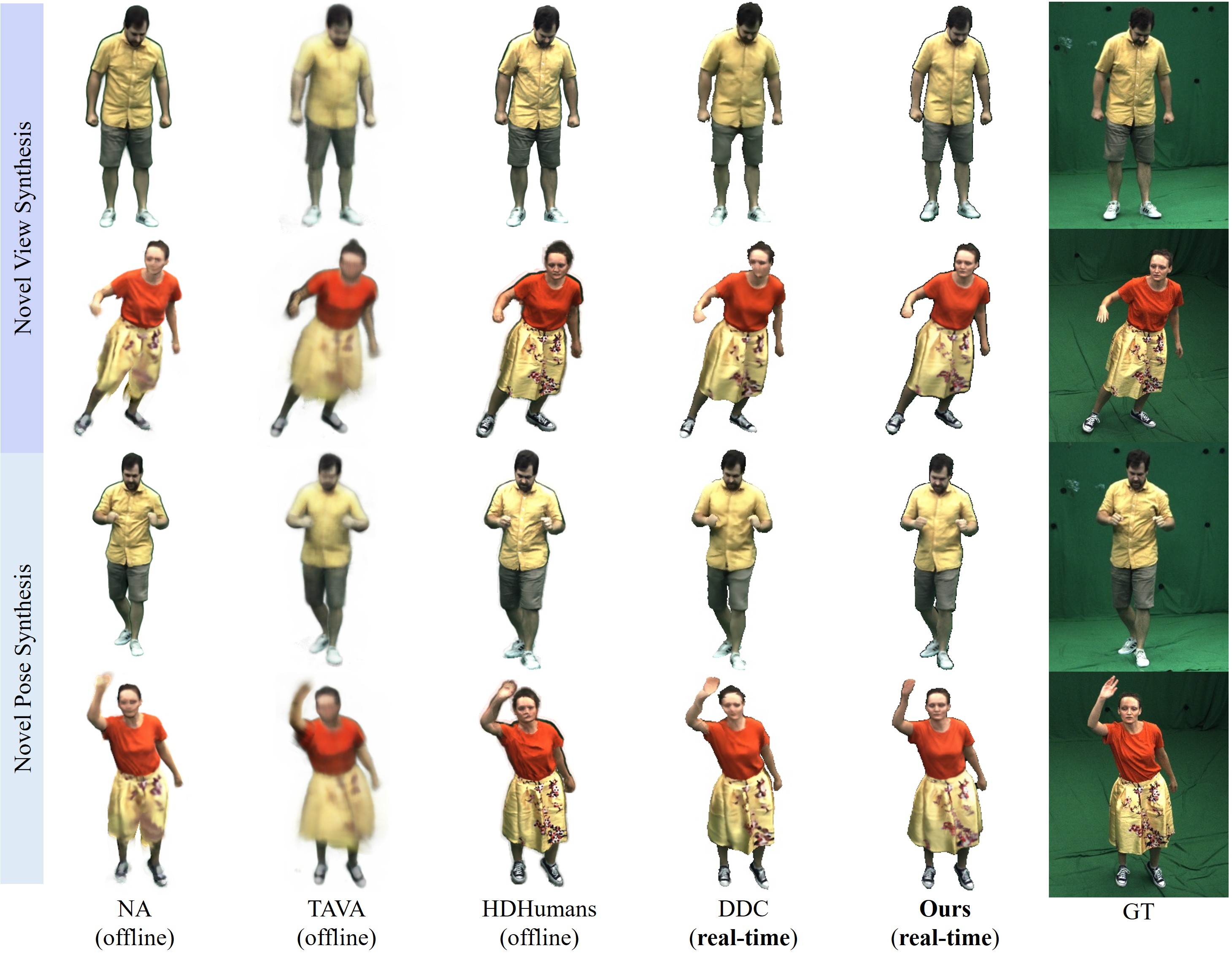}
\caption{
\textbf{Qualitative image synthesis comparison.}
Here, we qualitatively compare the image synthesis quality of our method and others~\cite{liu2021neural,li2022tava,habermann2022hdhumans,habermann2021real}.
Note that the visual quality of our method is better or comparable to current \textit{offline} approaches~\cite{liu2021neural,li2022tava,habermann2022hdhumans} while showing superior quality compared to other real-time methods~\cite{habermann2021real}.
}
\label{fig:qualcomp}
\end{figure*}
\begin{figure*}
\centering
\includegraphics[width=1.0\linewidth]{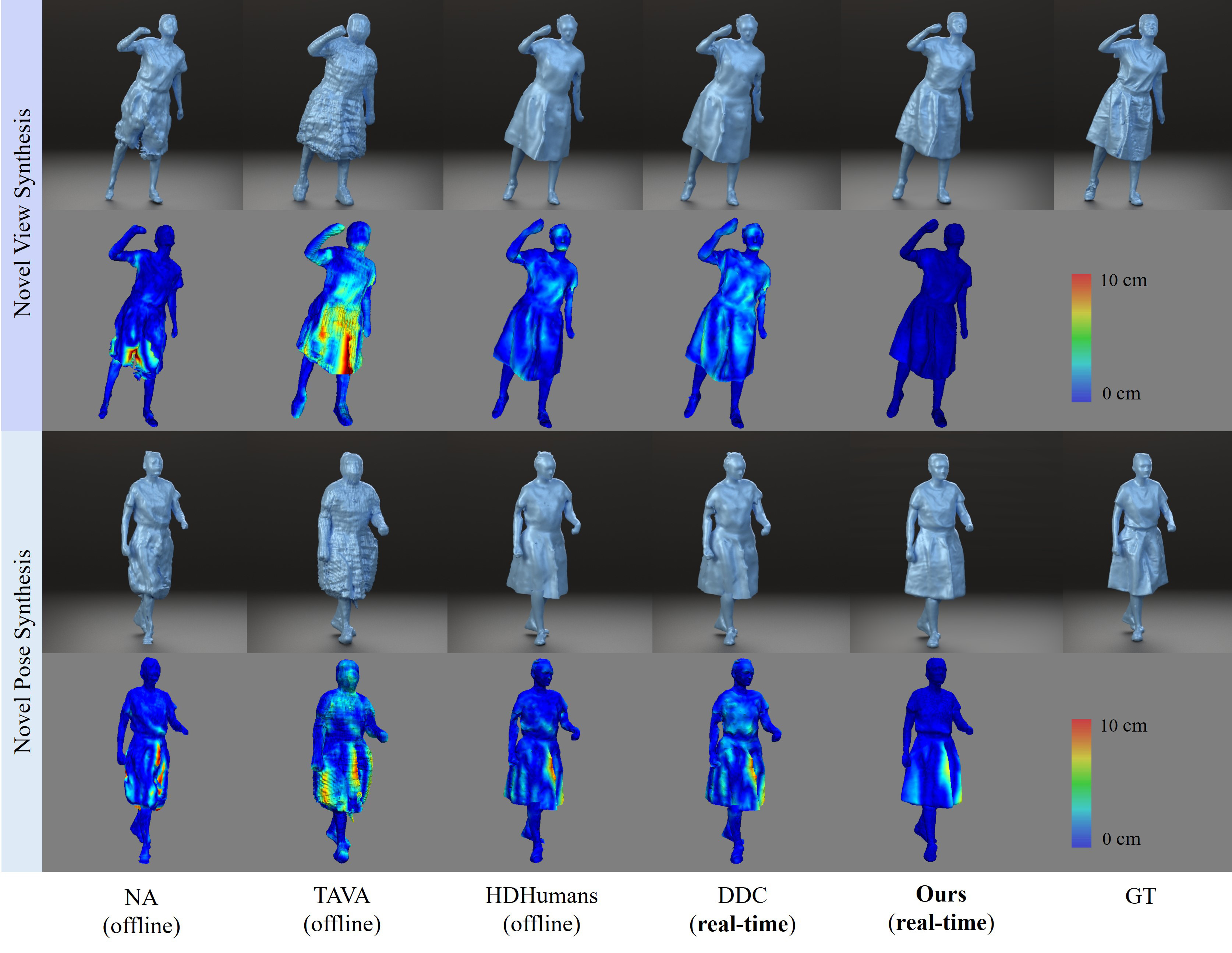}
\caption{
\textbf{Qualitative geometry comparison.}
Here, we qualitatively compare the generated geometry with other works~\cite{liu2021neural,li2022tava,habermann2022hdhumans,habermann2021real}.
Each row of the generated geometry is followed by its corresponding error map.
Note that our method achieves the highest geometric details while also achieving real-time performance.
This is consistent for both training and novel skeletal motions.
}
\label{fig:qualcompgeo}
\end{figure*}
%
%

\subsection{Comparisons} \label{subsec:comparisons}
\textbf{Competing Methods.}
We compare our method with two types of previous methods including 1) NA~\cite{liu2021neural} and TAVA~\cite{li2022tava}, which adopt a piece-wise rigid mapping with learned residual deformations, 2) \citet{habermann2022hdhumans} and DDC~\cite{habermann2021real}, which model surface deformation.
Note that only DDC supports real-time performance while other approaches require multiple seconds per frame.
We compare on two subjects from the third-party DynaCap dataset~\cite{habermann2021real}, one wearing a loose type of apparel, referred to as \textit{Loose Clothing}, and the other one wearing a tight type of apparel, referred to as \textit{Tight Clothing}.
\par 
\textbf{Metrics.}
In the following, we explain the individual metrics for quantitative comparisons. 
For assessing the quality of geometry, we provide measurements of the Chamfer distance, which computes the discrepancy between the pseudo ground truth obtained using an implicit surface reconstruction method~\cite{neus2} and the reconstructed shape results. 
A lower Chamfer distance means a closer alignment between two shapes, indicating a higher quality reconstruction.
We average the per-frame Chamfer distance over every 10th frame.
To evaluate the quality of image synthesis, we employ the widely-used Peak Signal-to-Noise Ratio (PSNR) metric. 
However, PSNR alone only captures the low-level error between images and has severe limitations when it comes to assessing the perceptual quality of images. 
Thus, PSNR may not accurately reflect the quality as perceived by the human eye. 
Consequently, we additionally report the learned perceptual image patch similarity (LPIPS) metric \cite{zhang2018unreasonable}, which is based on human perception.
We follow the test split from the DynaCap dataset having 4 test views. 
Here, metrics are averaged over every 10th frame and over all test cameras.
\par 
\textbf{Geometry.}
In Tab.~\ref{tab:comp_view_rec} and \ref{tab:comp_pose_geo}, we conduct a quantitative evaluation of our method and competing approaches to assess their performance in terms of geometry synthesis for training and test motions.
For NA and TAVA, we employed Marching Cubes to extract per-frame reconstructions from the learned NeRF representation. 
However, these recovered geometries exhibit a significant amount of noise due to the lack of geometry regularization and piece-wise rigid modeling during learning.
As a result, these methods demonstrate inferior performance compared to our approach, both, visually and quantitatively.
Compared to NA and TAVA, DDC yields better performance as it models the space-time coherent template deformation.
However, DDC relies solely on image-based supervision to learn the deformations, which only yields fixed wrinkles derived from the base template and struggles to track the dynamic wrinkle patterns.
In contrast, \cite{habermann2022hdhumans} outperforms DDC in the overall surface quality with the inclusion of the NeRF, while it falls short in real-time reconstruction.

Besides, we also qualitatively compare the generated geometry of our approach with other works as shown in Fig.~\ref{fig:qualcompgeo}. 
Note that our method achieves the highest geometric details among all methods while also achieving real-time performance. 
This is consistent for, both, training and novel skeletal motions.
We refer to the supplemental video to better see the dynamic deformations, which our method is able to recover.
\par 
\textbf{Novel View Synthesis.}
We quantitatively evaluate the novel view synthesis quality of different approaches as shown in Tab.~\ref{tab:comp_view_rec}.
For the comparison within real-time methods, our approach outperforms the competing method DDC~\cite{habermann2021real} by a substantial margin in terms of PSNR and LPIPS.
The difference in PSNR is relatively less pronounced, as it is less sensitive to blurry results and does not faithfully reflect the realism perceived by humans.
For the biased comparison with non real-time methods, our method still outperforms previous works remarkably in terms of PSNR, further verifying the effectiveness of our approach in achieving superior results.
The LPIPS score of our approach is inferior to HDHumans~\cite{habermann2022hdhumans}.
We speculate that their density-based formulation might help to achieve slightly better image quality compared to the SDF-based representation that we use.
Additionally, they have a significantly higher computational budget, which should also be considered here as their method runs multiple seconds per frame while we achieve real-time performance.
In summary, Tab.\ref{tab:comp_view_rec} provides quantitative confirmation of our method's outstanding view synthesis performance.
Even though the comparison is biased towards non-real time methods like NA, TAVA, and HDHumans~\cite{habermann2022hdhumans}, the overall superiority of our approach, including its PSNR performance reinforces the validity of our method.
\par 
We also qualitatively compare our approach with previous works in terms of the novel view synthesis.
As shown in Fig.~\ref{fig:qualcomp}, the visual quality of our method is better or on par with current offline approaches including NA, TAVA, and HDHumans~\cite{habermann2022hdhumans} while showing superior quality compared to the real-time method DDC.
Specifically, the view synthesis results of TAVA is very blurry and contains obvious visual artifacts as this method inherently struggles to handle more challenging datasets like ours and the DynaCap dataset with plentiful variations and long sequences.
NA shows reasonable performance on subjects wearing tight type of apparel. 
However, for loose clothing, it becomes obvious that their method cannot correctly handle the skirt region since the residual deformation network fails to correctly account for this.
The results of \citet{habermann2022hdhumans} are less blurry compared to the aforementioned methods and can compete with our method, however, at the cost of real-time performance.
DDC is capable of capturing medium frequency wrinkles well, but lacks finer details. 
In contrast to that, our method is able to achieve high-fidelity synthesis with the sharper details in real time.
\begin{table}[t]
\renewcommand\tabcolsep{10.0pt}
\small
    \centering
    \caption{\textbf{Ablation study}.
    We quantitatively evaluate our design choices for the novel view synthesis and geometry generation on a subject wearing a loose type of apparel.
    Note that our final design achieves the best quantitative results in all metrics.
    }
    \label{tab:ablation}
    \begin{tabular}{l|c|c|c}
    \hline
    \multicolumn{4}{c}{\textit{Training Poses (Loose Clothing)}} \\
    \hline
    Methods & \textbf{PSNR} $\uparrow$  & \textbf{LPIPS} $\downarrow$ & \textbf{Cham.} $\downarrow$ \\
    \hline
    w/ skin. mesh       & 27.82  & 30.36 & 3.768  \\
    w/o map opt.       & 30.21  & 29.20 & 1.714 \\
    w/ can. tri-plane   & 30.54  & 23.89 & 1.807   \\
    w/ MLP              & 30.09  & 25.47 & 2.008  \\
    2D Feat + D              & 31.12  & 23.92 & 1.521  \\
    w/o GMC SDF              & 31.57  & 16.71 & 1.532  \\
    w/o GMC               & 31.16  & 17.19 & 1.595  \\
    \hline
    \textbf{Ours}       & \textbf{31.68}  & \textbf{16.14} & \textbf{1.488}  \\
    \hline
    \end{tabular}
\end{table}

\begin{figure*}[!t]
\centering
\includegraphics[width=1.0\linewidth]{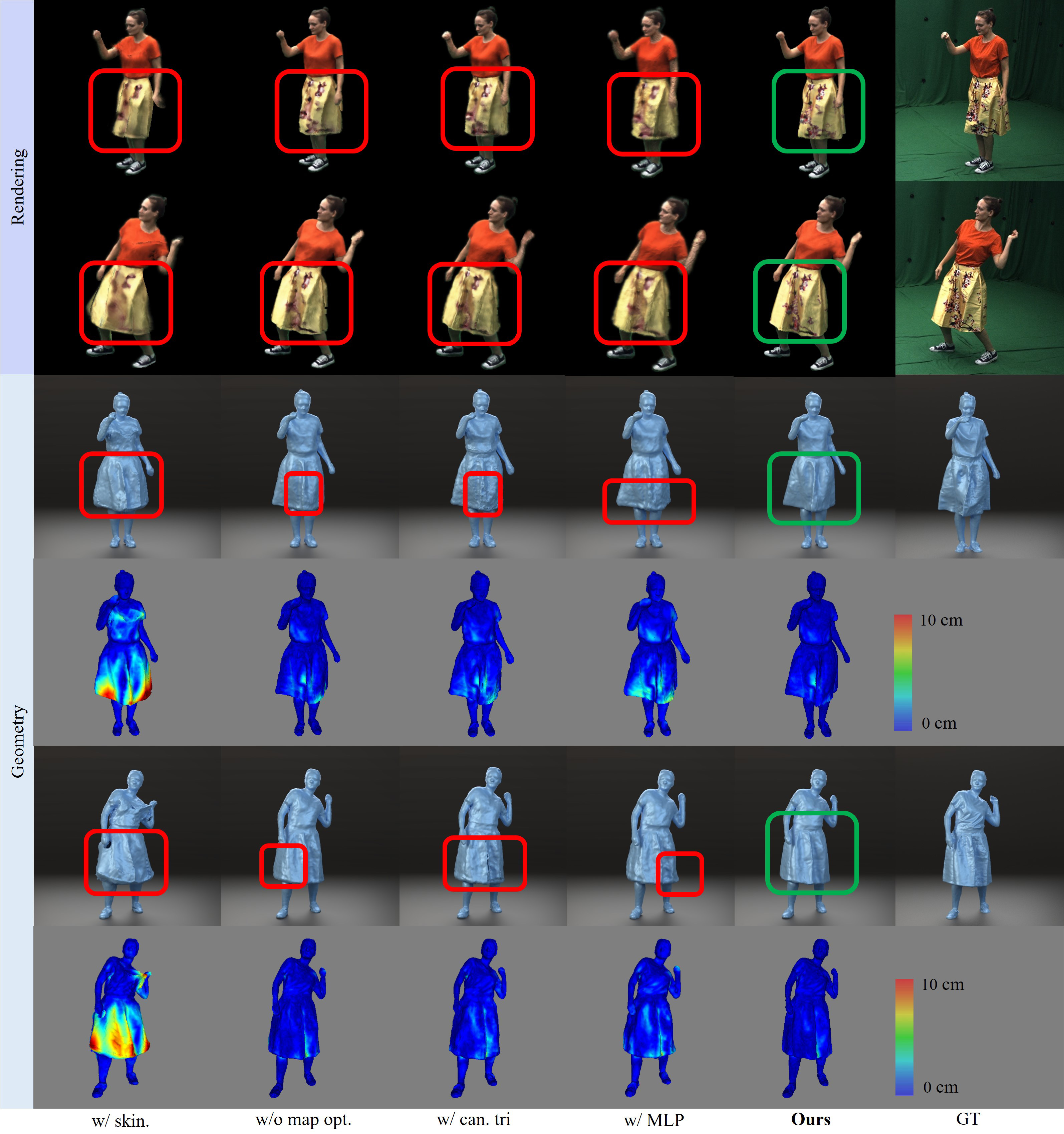}
\caption{
\textbf{Ablation study.}
We qualitatively evaluate our individual design choices for novel views in terms of image and geometry synthesis.
Note that each row of the generated geometry is followed by its corresponding error map.
Our results demonstrate that our proposed method consistently outperforms the baselines, which shows the superiority of our method.
}
\label{fig:qualablation}
\end{figure*}
\par 
\textbf{Novel Pose Synthesis.}
The same tendency can be observed when comparing to other works in terms of novel pose synthesis as shown in Tab.~\ref{tab:comp_pose_geo} and Fig.~\ref{fig:qualcomp}. 
Again, our method achieves the best perceptual results due to the high quality synthesis of our approach.
In terms of PSNR, some methods such as NA achieve a good score although results are notably very blurred or not photorealistic.
\cite{habermann2022hdhumans} still achieves the best LPIPS score while it is limited to non-real time synthesis.
With the same real-time configuration, our method clearly outperforms previous work of DDC significantly in terms of synthesis quality.
%
%
\subsection{Ablation Studies} \label{subsec:ablation}
We quantitatively ablate our design choices on the novel view synthesis task in Tab.~\ref{tab:ablation}.
A qualitative ablation study is also performed for novel views in terms of image and geometry, as shown in Fig.~\ref{fig:qualablation}. 
For an ablation on the novel pose task, we refer to the supplemental material.
\par 
In the following, we first compare our design choices of using non-rigid space canonicalization and our proposed UTTS space to alternative baselines.
\par 
\textbf{Skinning-based Deformation Only.}
As shown in Tab.~\ref{tab:ablation}, employing pure skinning-based deformation mentioned on template mesh, i.e. setting $\mathbf{Y}_v=\mathbf{M}_v$ in Eq.~\ref{eq:ddc_deform}, without a non-rigid residue (i.e., \textbf{w/ skin. mesh}) leads to significant performance degradation in terms of synthesis quality (evaluated by PSNR and LPIPS) and geometry quality (evaluated by Chamfer loss).
The qualitative results in Fig.~\ref{fig:qualablation} also validate the performance drop with pure skinning-based deformation.
The reason is that the mapping into UTTS is less accurate since skinning alone cannot account for non-rigidly deforming cloth areas leading to mapping collisions and wrongly mapped points.
This confirms that our design choice of accounting for non-rigid deformations within the mapping procedure via a deformable model, which is also gradually refined throughout the training, is superior over piece-wise rigid, i.e. skinning-based only transformations.
\par 
\textbf{SDF-driven Surface Refinement.}
Next, we evaluate the impact of our second training phase, where we update the learnable parameters of the deformable human model to better fit the SDF (see SDF-driven Surface Refinement in Sec.~\ref{subsec:supervision}).
Our SDF-driven surface refinement is beneficial to both synthesis quality and geometry quality as evident from the ablation study.
As mentioned earlier, the better the deformable model approximates the true surface, the smaller the distance $d_\mathrm{max}$ (see discussion at the end of Sec.~\ref{subsec:mapping}) can become reducing the cases of points mapping onto edges and vertices.
Discarding SDF-driven surface refinement (i.e., \textbf{w/o map opt.}) will result in a performance drop as shown in Tab.~\ref{tab:ablation} and Fig.~\ref{fig:qualablation}.
\par 
\textbf{Tri-plane in Canonical Pose Space.}
Next, we evaluate the design of our UTTS space, which suggests mapping global points into a 3D texture space. 
A popular alternative in literature is the \textit{canonical unposed} 3D space, i.e., the character is usually in a T-pose.
For this ablation (referred to as \textbf{w/ can. tri-plane}), we, therefore, placed the tri-plane into this canonical unposed 3D space and evaluated the performance. 
We found that there are severe mapping collisions, particularly in wrinkled clothing areas and, thus, this mapping showed a decrease in performance (see Tab.~\ref{tab:ablation} and Fig.~\ref{fig:qualablation}).
In consequence, our mapping into UTTS space is preferable.
\par 
Next, we compare our motion-dependent tri-plane encoder against several baselines to evaluate its 
effectiveness. 
\par 
\textbf{MLP-only.}
To evaluate the importance of our tri-planar motion encoder (see Sec.~\ref{subsec:tri_plane}), we compare to a pure MLP-based representation (i.e., \textbf{w/ MLP}).
Here, we remove the tri-plane encoding and instead feed the skeletal pose directly into the MLP, 
It can be clearly seen that this design falls short in terms of visual quality and geometry recovery Tab.~\ref{tab:ablation} and Fig.~\ref{fig:qualablation}.
This can be explained by the fact that the representation capability of the MLP is insufficient to model the challenging dynamic human body and clothing.
A deeper MLP-based architecture could help here, however, at the cost of real-time performance since deeper architectures are significantly slower as the MLP has to be evaluated for \textit{every} sample along \textit{every} ray.

\par 
\textbf{2D Feature and Pose-encoded D.} To assess the necessity of the triplane for our task, we orchestrate an ablation experiment, which replaces the motion-dependent triplane features with 2D features and positionally encoded distances, termed as \textbf{2D Feat + D}. To achieve this, we adopt the original UNet architecture for generating 2D features from motion textures. Notably, similar to our 3D-aware convolutional motion encoder, the global motion code is channel-wise concatenated to the bottleneck feature maps. As illustrated in Tab.~\ref{tab:ablation} and Fig.~\ref{fig:qualablation}, our motion-dependent triplane representation exhibits superior appearance/geometry accuracy because the d-dimension in our motion-dependent triplane can encode motion-aware features by indexing respective feature planes (UD/VD), while \textbf{2D Feat + D} only allows the UV-plane to be motion-aware. 

\par 
\textbf{Global Motion Code.}
We conduct two ablations to demonstrate the effectiveness of the global motion code. The first ablation removes the global motion code from the SDF MLP input features, referred to as \textbf{w/o GMC SDF}. Moreover, we conducted the second ablation that eliminates the global motion code from both the triplane bottleneck features and the SDF MLP input features, termed as \textbf{w/o GMC}. The results in Tab.~\ref{tab:ablation} and Fig.~\ref{fig:qualablation} indicate that removing global motion code from the SDF (\textbf{w/o GMC SDF}) leads to a minor drop in the performance, while removing global motion code from the triplane bottleneck and SDF (\textbf{w/o GMC}) experiences a more significant drop due to the lack of global motion awareness.

%
%
\section{Applications}
In this section, we will introduce two applications built upon TriHuman: the TriHuman Viewer, a real-time interactive system designed for inspecting and generating highly detailed clothed humans (Sec.~\ref{sec:system}), and the consistent texture editing supported by TriHuman (Sec.~\ref{sec:application}).
%
%

\subsection{TriHuman Viewer} \label{sec:system}

\begin{figure}[t]
\includegraphics[width=1.0\linewidth]{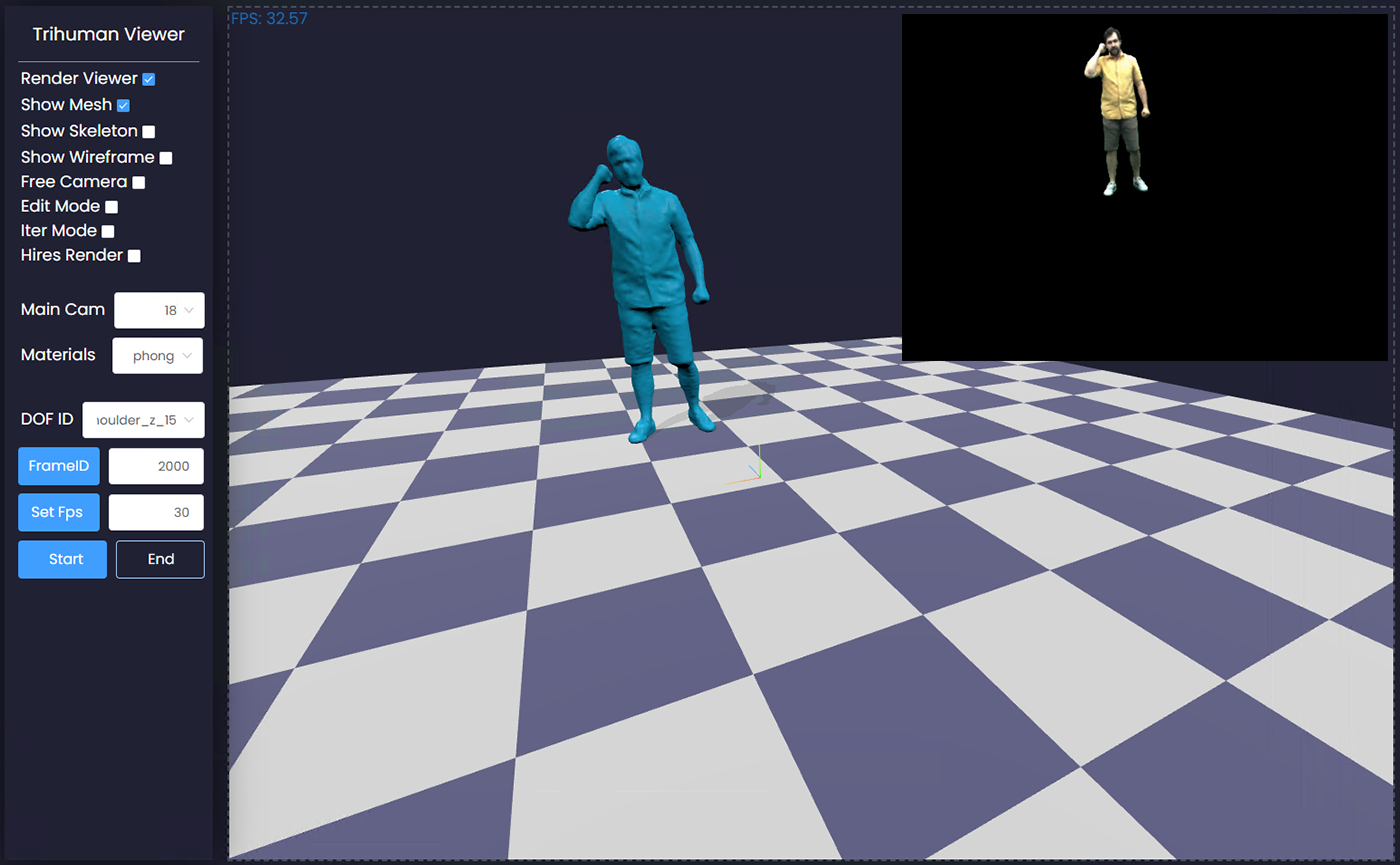}
  \caption{
    The TriHuman Viewer offers a real-time interface that enables users to examine the rendering and geometry of training and validation motions. Furthermore, Trihuman empowers users to customize camera positions and skeletal DOFs for creating novel-view renderings and novel motion geometries and renderings.
  }
  \label{fig:interface}
\end{figure}

Building upon TriHuman, we introduce an interactive real-time system, i.e., the TriHuman Viewer (Fig.~\ref{fig:interface}), that enables users to inspect and generate high-fidelity clothed human geometry and renderings, given skeletal motion and camera poses as inputs. We refer to the supplemental video and document for more details regarding the supported interactions and the runtime for each algorithmic component.
%
%
\subsection{Texture Editing}
\label{sec:application}
As highlighted in Sec.~\ref{sec:overview}, TriHuman can generate detailed geometry with consistent triangulation, opening up new possibilities for a broad spectrum of downstream applications. Here, we use consistent texture editing as an illustrative example of such applications.
\par
\begin{figure}[t]
\includegraphics[width=1.0\linewidth]{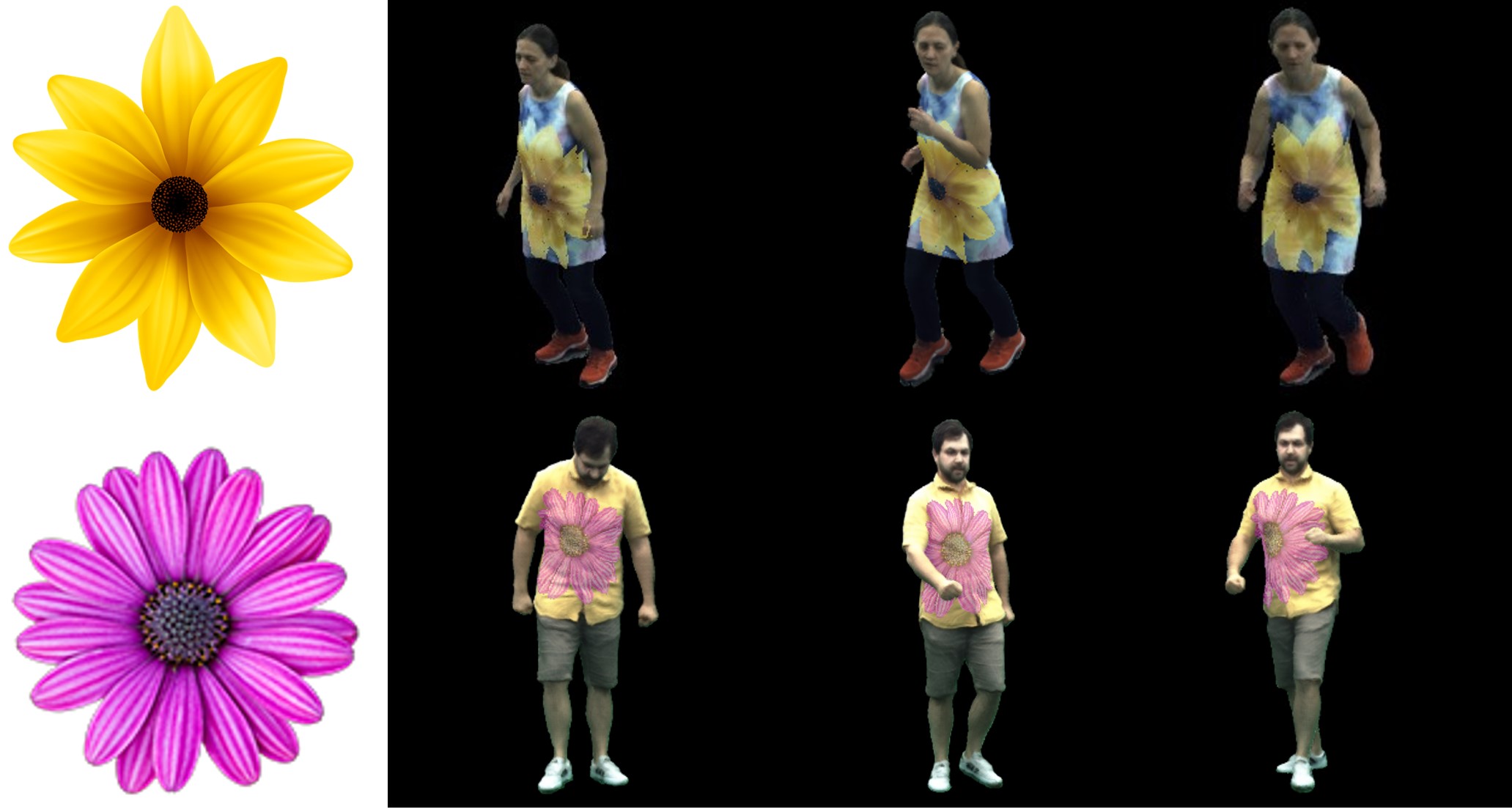}
  \caption{
    The results for consistent texture editing. The flowers in the leftmost column can be seamlessly integrated into the clothed human rendering, faithfully adapting to the clothed human's deformations and preserving occlusions caused by various poses.
  }
  \label{fig:supplapplication}
\end{figure}

Fig.~\ref{fig:supplapplication} presents the results for consistent texture editing, which can be achieved through the following steps:
Firstly, we select an image with an alpha channel, serving as the edits to the texture map for the character's template mesh.
Next, we render the texture color and the alpha value through the rasterization process applied to the textured template mesh.
Finally, we achieve consistent texture editing results by alpha-blending the neural-rendered character imagery with the rasterized texture.
Notably, thanks to the high-fidelity and consistent geometry generated by TriHuman, the rendered edits follow the wrinkle deformation of the clothing. Moreover, the edited result effectively retains the occlusions resulting from different poses.
%
%
\section{Limitations and Future Work} 
\label{sec:future}

While our approach enables controllable, high-quality, and real-time synthesis of human appearance and geometry, there are some limitations, which we hope seeing addressed in the future.
\par 
First, our model is currently not capable of generating re-lightable human appearance since we are not decomposing appearance into material and lighting.
However, since the geometry reconstructed by our method is highly accurate, it becomes possible to incorporate re-lightability into our model to enhance the realism and visual coherence of the reconstructed human body in various applications and environments.
Second, we are currently representing the human surface as an SDF and explicit mesh model.
However, for the hair region, such a representation might not be ideal. 
Future work could consider a hybrid density and SDF-based representation accounting for the different body parts and regions that may prefer one representation over the other.
Third, our model does not support generalization across identities, which is also the limitation of most models for detailed human reconstruction. 
A possible avenue for future work could be to leverage transfer learning approaches, where pre-trained models on large-scale datasets are fine-tuned or adapted to specific identities.

Moreover, our method does not support generating controllable facial expression rendering due to the absence of facial tracking in our dataset, which could be addressed in the future by incorporating facial tracking into the dataset. Furthermore, like all existing methods, our method cannot model surface dynamics induced by external forces like wind. A promising future direction would be introducing the physical constraints into the training of the geometry and appearance generation models.

%
%
\section{Conclusion} \label{sec:conclusions}
We introduced TriHuman, a novel approach for controllable, real-time, and high-fidelity synthesis of space-time coherent geometry and appearance solely learned from multi-view video data.
Our method excels in reconstructing and generating a virtual human with challenging loose clothing of exceptional quality.
The key ingredient of our approach lies in a deformable and pose-dependent tri-plane representation, which enables real-time yet superior performance.
A differentiable and mesh-based mapping function is introduced to reduce the ambiguity for the transformation from global space to canonical space.
The results on our new benchmark dataset with challenging motions unequivocally demonstrate significant progress towards achieving more lifelike and higher-resolution digital avatars, which hold great importance in the emerging realms of virtual reality (VR).
We anticipate that the proposed model with the new benchmark datasets can serve as a robust foundation for future research.


\bibliographystyle{ACM-Reference-Format}
\bibliography{reference}

\setcounter{figure}{0}
\setcounter{table}{0}
\setcounter{equation}{0}
\appendix

\section{Overview}
In this supplemental material, we provide more details regarding the following aspects: 
In Sec. ~\ref{sec:supplloss}, we delve into the implementation details of loss terms for training. 
In Sec. ~\ref{sec:supplmap}, we demonstrate the effectiveness of the proposed deformable triplane and the UVD mapping paradigm in reducing mapping ambiguity. In Sec. ~\ref{sec:supplnetwork}, we depict the network architectures and the hyperparameters for the trainable components of TriHuman. In Sec. ~\ref{sec:supplablation}, we present more ablation studies to assess the design choices of our model. Finally, in Sec. ~\ref{supplsec:system}, we elaborate the real-time interactive system, i.e., TriHuman Viewer, and analyze the runtime for each component.

%
%
\section{Loss Terms} \label{sec:supplloss}
In this section, we provide more implementation details regarding the loss terms introduced in the main paper. 
%
%
\subsection{Seam Loss}
In the main paper, we parameterize the motion-dependent clothed human surface with a deformable UVD texture cube, i.e., Undeformed Tri-plane Texture Space (UTTS). 
While UTTS provides high-quality geometry and appearance, the seams on the UV paradigm may lead to gaps in the posed geometry due to discontinuities of the features on either side of the UV seam. 
To address this issue, we propose a UV seam loss $\mathcal{L}_\mathrm{seam}$ that penalizes the discontinuity of the implicit geometry near the UV seams:
%
\begin{equation}
\begin{split}
s_{\mathrm{a},i,f}, \mathbf{q}_{\mathrm{a},i,f} &= \mathcal{H}_\mathrm{sdf}(\mathbf{F}_{\mathrm{a},i,f}, \mathbf{g}_f, p(\bar{\mathbf{x}}_{\mathrm{a},i}); \Gamma) \\
s_{\mathrm{b},i,f}, \mathbf{q}_{\mathrm{b},i,f} &= \mathcal{H}_\mathrm{sdf}(\mathbf{F}_{\mathrm{b},i,f}, \mathbf{g}_f, p(\bar{\mathbf{x}}_{\mathrm{b},i}); \Gamma) \\ 
\mathcal{L}_\mathrm{seam} &= \frac{1}{S}\sum_{i=1}^S \| s_{\mathrm{a},i,f} - s_{\mathrm{b},i,f} \|_2
\end{split}
\end{equation}
%
where $\bar{\mathbf{x}}_{\mathrm{a},i}$ and $\bar{\mathbf{x}}_{\mathrm{b},i}$ denotes the sample pairs near the corresponding seam edges in the UV space and $S$ is the number of seam samples. 
$s_{\mathrm{a},i,f}$ and $s_{\mathrm{b},i,f}$ denotes the SDF value computed at the sampled position. Generating the paired samples in the UVD texture volume $\bar{\mathbf{x}}_{\mathrm{a},i}$, $\bar{\mathbf{x}}_{\mathrm{b},i}$ for evaluating the seam loss is achieved through:
%
\begin{equation}
\begin{split}
\bar{\mathbf{x}}_{\mathrm{a},i} &= \mathbf{p}_{\mathrm{st},\mathrm{a},j} + \alpha_{\mathrm{seam},i} * \mathbf{r}_{\mathrm{a},j} + \epsilon_{\mathrm{seam},i} * \mathbf{n}_{\mathrm{a},j} + h_{i} \\
\bar{\mathbf{x}}_{\mathrm{b},i} &= 
\mathbf{p}_{\mathrm{st},\mathrm{b},j} + \alpha_{\mathrm{seam},i} * \mathbf{r}_{\mathrm{b},j} + \epsilon_{\mathrm{seam},i} * \mathbf{n}_{\mathrm{b},j} + h_{i}
\end{split}
\end{equation}
%
where $\mathbf{p}_{\mathrm{st},\mathrm{a},j}$ and $\mathbf{p}_{\mathrm{st},\mathrm{b},j}$ denotes the startpoint of the selected seam edge pairs; 
$\alpha_{\mathrm{seam},i}$ indicates the linear interpolation factor for the sampling on the seam edge pairs; $\mathbf{r}_{\mathrm{a},j}$ and $\mathbf{r}_{\mathrm{b},j}$ denotes the oriented length for the seam edges; 
$\epsilon_{\mathrm{seam},i}$ denotes the random offsets along the selected seam edge pair normals $\mathbf{n}_{\mathrm{a},j}$ and $\mathbf{n}_{\mathrm{b},j}$;$h_{i}$ indicates the random offset along the height dimension, which follows the uniform distribution on the interval $[-0.05, 0.05]$. 
During training, $\epsilon_{\mathrm{seam},i}$ is set to $0.01$ empirically.
%
%
\subsection{SDF Loss}
In the SDF-driven surface refinement stage, we adopt an SDF loss $\mathcal{L}_\mathrm{sdf}$ that guides the explicit template to fit the detailed implicit surface by forcing the SDF value of the posed template mesh vertices $\mathbf{v}^{\prime}_{i}$ to be zero:
%
\begin{equation}
\begin{split}
s_{i,f}, \mathbf{q}_{i,f} &= \mathcal{H}_\mathrm{sdf}(\mathbf{F}_{i,f}, \mathbf{g}_f, p(\bar{\mathbf{v}}^{\prime}_{i}); \Gamma) \\
\mathcal{L}_\mathrm{sdf} &= \frac{1}{N}\sum_{i=1}^N\| s_{i,f} \|_2
\end{split}
\end{equation}
where $\bar{\mathbf{v}}^{\prime}_{i}$ indicates the canonicalized vertex position of the updated posed template mesh. 
Notably, for better training stability, we only optimize the explicit template while fixing the weights for the detailed implicit field.
%
%
\subsection{Surface Regularization Loss Term}
In the SDF-driven surface refinement stage, we propose to adopt multiple surface regularization terms to maintain the overall smoothness of the updated template mesh while not losing the high-frequency geometry details. 
\par 
\noindent \textbf{Laplacian Loss}.
To avoid artifacts due to the inconsistent local deformations, we adopt the Laplacian loss $\mathcal{L}_\mathrm{reg}$ as the geometry regularizer, which penalizes the differences between the Laplacians of the SDF-updated and the original template mesh
%
\begin{equation}
\mathcal{L}_\mathrm{reg} = \frac{1}{\mathbf{N}} \sum_{i=1}^{N} \|(\mathbf{L} \mathcal{V}(\boldsymbol{\theta}_{f':f};\Omega))_i - (\mathbf{L} \mathcal{V}(\boldsymbol{\theta}_{f':f};\Omega^{\prime}))_i\|_{2}
\end{equation}
%
where $\mathbf{L}$ indicates the mesh Laplacian matrix; $\mathcal{V}(\boldsymbol{\theta}_{f':f};\Omega^{\prime})$ denotes the SDF-updated template mesh.
\par 
\noindent \textbf{Surface Smoothing Loss}.
To preserve the overall smoothness of the updated template mesh, we employ a Laplacian-based smoothing term $\mathcal{L}_\mathrm{zero}$ that pushes the Laplacian computed from the deformed template to zero
%
\begin{equation}
\mathcal{L}_\mathrm{reg} = \frac{1}{\mathbf{N}} \sum_{i=1}^{N} \|(\mathbf{L} \mathcal{V}(\boldsymbol{\theta}_{f':f};\Omega^{\prime}))_i\|_{2}.
\end{equation}
%
\par 
\noindent \textbf{Normal Consistency Loss}. 
As the flipping faces on the template meshes may lead to abrupt changes in the space mapping, we adopt a normal consistency loss $\mathcal{L}_\mathrm{normal}$ that prevents flipping faces by penalizing the discrepancies in face normals among adjacent faces
%
\begin{equation}
\mathcal{L}_\mathrm{normal} = \frac{1}{\mathrm{N}_{\mathrm{f}}}(\sum_{i=1}^{\mathrm{N}_{\mathrm{f}}}\frac{\sum_{j=1}^{\mathrm{N}_{\mathrm{f},i}}(1 - \mathbf{n}_{\mathrm{f},i} \cdot \mathbf{n}_{\mathrm{f},i,j})}{\mathrm{N}_{\mathrm{f},i}})
\end{equation}
%
where $\mathrm{N}_{\mathrm{f}}$ indicates the number for faces of the template mesh; $\mathrm{N}_{\mathrm{f},i}$ denotes the number of faces adjacent to face $i$; $\mathbf{n}_{\mathrm{f},i,j}$ refers to the face normals of the adjacent faces of face $i$.
\par 
\noindent \textbf{Face Stretching Loss}. 
As degraded faces would lead to numerical errors for the space mapping, we adopt the face stretching loss $\mathcal{L}_\mathrm{area}$ that reduces overly stretched faces by minimizing the deviation of the edge lengths within each face for the deformed template $\mathcal{V}(\boldsymbol{\theta}_{f':f};\Omega^{\prime})$
%
\begin{equation} 
\mathcal{L}_\mathrm{area} = \frac{1}{\mathrm{N}_{\mathrm{f},i}}\sum_{i=1}^{\mathrm{N}_{\mathrm{f}}}\mathrm{Var}(e_{\mathrm{0},i},e_{\mathrm{1},i},e_{\mathrm{2},i})
\end{equation}
%
where $e_{\mathrm{0},i},e_{\mathrm{1},i},e_{\mathrm{2},i}$ denote the edges for the $i$th face of the template mesh. 
%
%
\section{Mapping Ambiguity} \label{sec:supplmap}
\begin{figure}[t]
  \includegraphics[width=0.95\linewidth]{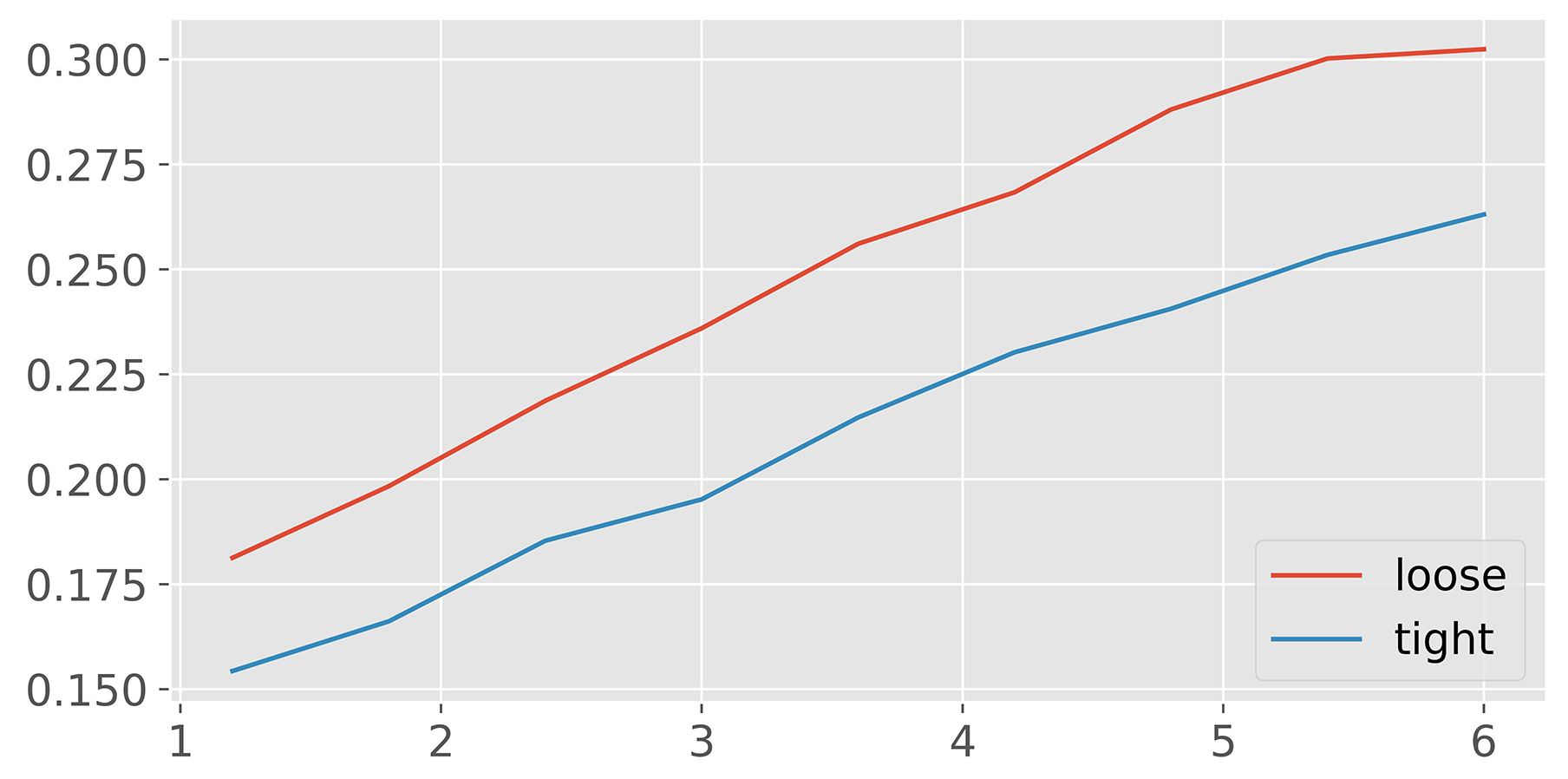}
  \caption{
    The mapping ambiguity concerning the height $d_\mathrm{max}$ of the Undeformed Tri-plane Texture Space (UTTS), tested on mesh sequences of subjects with loose and tight outfits. The X-axis denotes the UTTS height in centimeters, while the Y-axis indicates the ratio of ambiguously-mapped samples. 
  }
  \label{fig:supplmapping}
\end{figure}

Fig. \ref{fig:supplmapping} illustrates the mapping ambiguity concerning the height of the Undeformed Tri-plane Texture Space (UTTS) on sequences of subjects with loose and tight clothing. 
Specifically, we recorded spatial samples for volume rendering to assess the mapping ambiguity and filter out the samples that do not fall within the bounds of UTTS. 
Notably, the ratio of ambiguity mapping significantly rises when increasing the UTTS height $d_\mathrm{max}$. 
This observation serves as evidence that we can reduce mapping ambiguity by updating the template mesh and decreasing the height of the UTTS. 
%
%
\section{Network Architecture} \label{sec:supplnetwork}
In this section, we provide more details regarding the network structures for each trainable component of TriHuman, namely, the motion-dependent deformable human model, the global motion encoder, the motion-dependent triplane generator, and the unbiased volume renderer.
%
%
\subsection{Motion-dependent Deformable Human Model}
In the main paper, we leverage graph-convolutional architecture proposed by~\cite{habermann2021real}, e.g., EGNet and DeltaNet, for modeling the coarse explicit geometry for the dynamic clothed humans. 
Moreover, we exclude the TexNet mentioned in ~\cite{habermann2021real} as we instead employ the Undeformed Tri-plane Texture Space (UTTS) to model the detailed appearance and geometry. 
%
%
\subsection{Global Motion Encoder}
In the main paper, the global motion code is fed to the bottleneck of the motion-dependent triplane generator and the SDF network to provide an awareness of global skeletal motion.
A tiny MLP with two hidden layers and a width of 128 is employed to generate the global motion code. 
The network takes the sliding window of the skeletal pose of the preceding three frames as input and produces a 16-channel global motion descriptor. 
Specifically, we factor out the root translation from the input skeletal pose DOFs.
%
%
\subsection{Motion-dependent Triplane Generator}
\begin{figure*}
  \includegraphics[width=0.90\linewidth,]{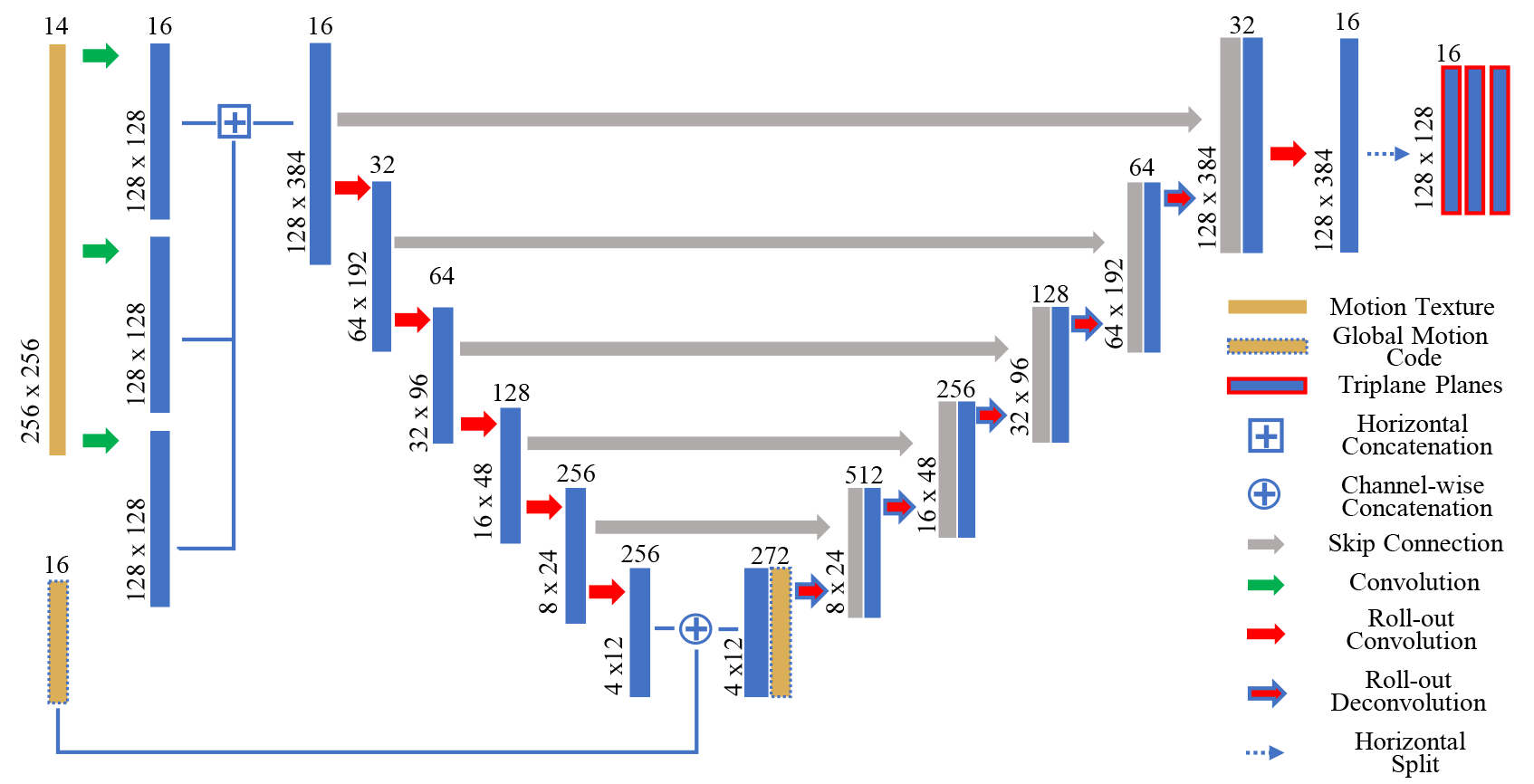}
  \caption{
    The network structure for the motion-dependent triplane generator. The network takes the motion texture rendered from the posed explicit template together with the global motion code as input and outputs a motion-aware triplane. 
  }
  \label{fig:suppltriplane}
\end{figure*}

Fig.~\ref{fig:suppltriplane} shows the network structure for the motion-dependent triplane generator. 
The network takes the motion texture rendered from the posed template and the global motion code as input and generates a motion-dependent triplane. 
To enhance the spatial contiguity of the triplane, we adopt roll-out convolution ~\cite{wang2022rodin} for fusing features across the triplane planes. 
Additionally, the global motion code is channel-wise concatenated with the bottleneck features of the triplane generator, providing awareness of global skeletal motion.

%
%
\subsection{Unbiased Volume Renderer}
Inspired by ~\citet{wang2021neus}, we adopt unbiased volume rendering to train our geometry and appearance model. 
However, the rendering formulation proposed in ~\cite{wang2021neus} was designed to model the appearance and geometry for static scenes.
Moreover, the large MLP network adopted leads to slow rendering and reconstruction. 
To achieve real-time, motion-aware appearance and geometry generation, we propose the following modifications to the original, unbiased volume renderer:
\par 
\noindent \textbf{Geometry Network}. 
We leverage an MLP for decoding motion-dependent geometry.
For each sample in the observation space, our proposed geometry network takes its counterpart in the Undeformed Tri-plane Texture Space(UTTS), tri-linear interpolated features from the motion-dependent triplane, together with the global motion code as input, and generates the SDF value and the shape features. 
In practice, we set the network width and the number of layers of the geometry MLP as 4 and 256 to strike a balance between quality and efficiency. 
\par 
\noindent \textbf{Appearance Network}. 
We adopt a 3-layer MLP with a width of 256 to model the motion-dependent appearance. 
The network takes the sample position in the observation space, the shape features from the geometry network, the ray direction, and the implicit surface normal as input and produces the color. 

%
%
\section{Ablation Studies} \label{sec:supplablation}
In this section, we provide more ablation studies to justify the design choices of TriHuman, namely, results on the testing pose, robustness against fewer cameras, and the height of UTTS.
%
%
\subsection{Ablation on Testing Poses}
Tab.~\ref{tab:supplablation} presents the quantitative comparison between our final model and the counterparts that adopt different design choices on the testing sequence of our dataset. 
The results confirm that our method has better generalization ability when dealing with novel motions than models with alternative design choices. 
\begin{table}[t]
\renewcommand\tabcolsep{10.0pt}
\small
    \centering
    \caption{\textbf{Ablation study}.
    We quantitatively evaluate our design choices for the novel motion appearance and geometry generation on a subject wearing a loose type of apparel.
    Note that our final design achieves the best quantitative results in all metrics.
    }
    \label{tab:supplablation}
    \begin{tabular}{l|c|c|c}
    \hline
    \multicolumn{4}{c}{\textit{Testing Poses (Loose Clothing)}} \\
    \hline
    Methods & \textbf{PSNR} $\uparrow$  & \textbf{LPIPS} $\downarrow$ & \textbf{Cham.} $\downarrow$ \\
    \hline
    w/ skin. mesh       & 27.20  & 30.17 & 4.187  \\
    w/o map opt.       & 27.72  & 26.59 & 2.845 \\
    w/ can. tri-plane   & 27.71  & 27.25 & 2.933   \\
    w/ MLP              & 27.04  & 27.16 & 3.022 \\
    2D Feat + D              & 27.25  & 23.26 & 1.631  \\
    w/o GMC SDF              & 27.68  & 22.95 & 2.783  \\
    w/o GMC               & 27.61  & 23.28 & 2.800  \\
    \hline
    \textbf{Ours}       & \textbf{27.78}  & \textbf{22.65} & \textbf{2.743}  \\
    
    \hline
    \end{tabular}
\end{table}

%
%
\subsection{Robustness Against Fewer Cameras}
\begin{figure}[h]
\centering
\includegraphics[width=0.99\linewidth]{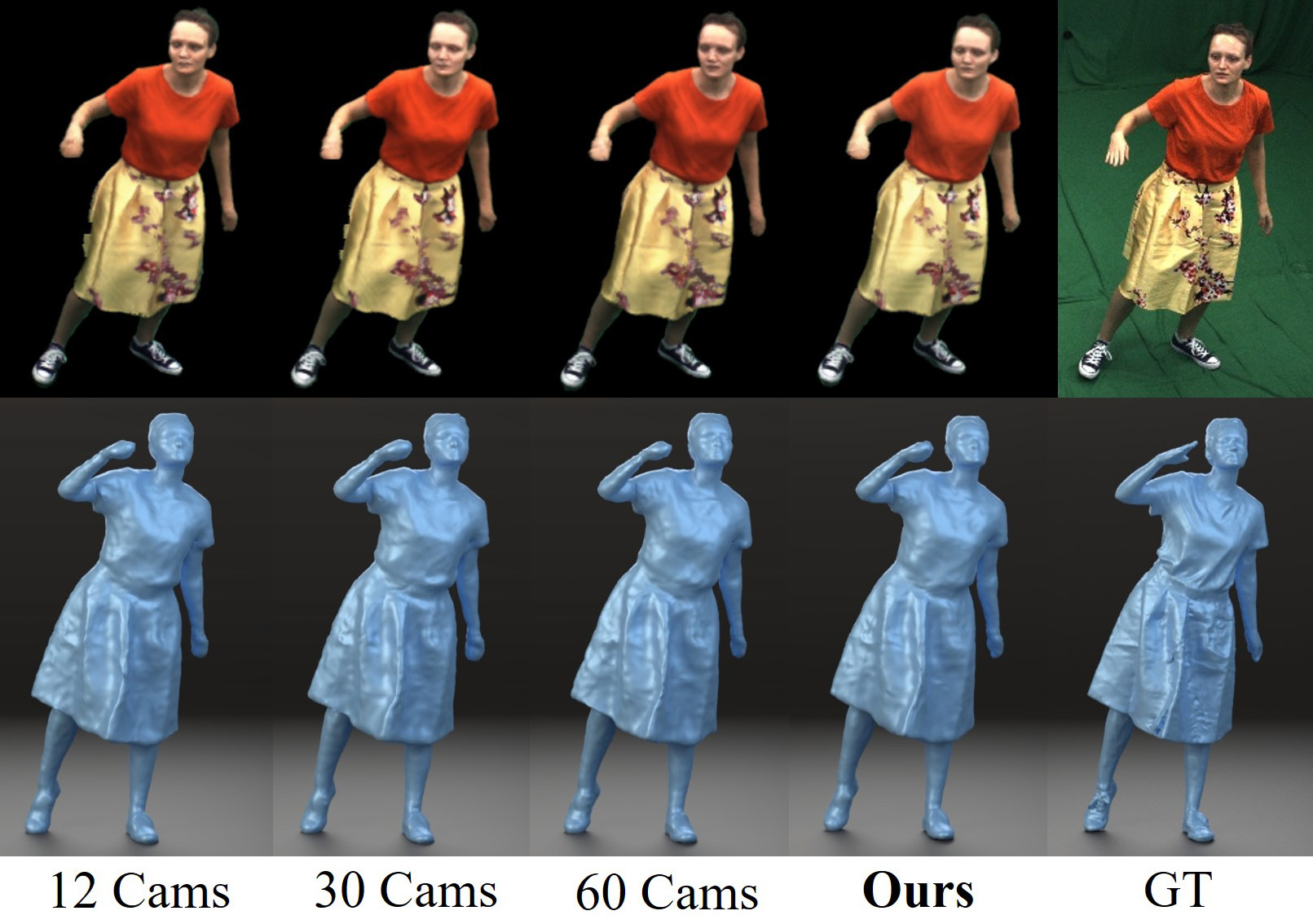}
\caption{
\textbf{Robustness Against Fewer Cameras.}
The rendering and geometry generated with our model trained with less input views.
Our results demonstrate that our proposed method is robust against less input views for training.
}
\label{fig:supplablationcam}
\end{figure}

To assess the robustness of our model against fewer input cameras, we conduct ablation experiments using videos captured from fewer camera views during training. 
As illustrated in Tab. \ref{tab:supplcamera} and Fig. \ref{fig:supplablationcam}, our method still achieves accurate results in view and geometry synthesis even with fewer input cameras.
\begin{table}[t]
\renewcommand\tabcolsep{10.0pt}
\small
    \centering
    \caption{\textbf{Robustness Against Fewer Cameras}.
    We quantitatively evaluate our model on the robustness against fewer cameras. 
    Notably, even with fewer cameras, our model performs well in view and geometry synthesis tasks.
    }
    \label{tab:supplcamera}
    \begin{tabular}{l|c|c|c}
    \hline
    \multicolumn{4}{c}{\textit{Training Poses (Loose Clothing)}} \\
    \hline
    Methods & \textbf{PSNR} $\uparrow$  & \textbf{LPIPS} $\downarrow$ & \textbf{Cham.} $\downarrow$ \\
    \hline
    12 Cameras      & 30.02  & 18.31 & 1.578  \\
    30 Cameras       & 31.54  & 17.26 & 1.558 \\
    60 Cameras       & 31.57  & 17.20 & 1.524 \\
    \hline
    \textbf{Ours}       & \textbf{31.68}  & \textbf{16.14} & \textbf{1.488}  \\
    \hline
    \end{tabular}
\end{table}

%
%
\subsection{Height of UTTS}
\begin{figure}[h]
\centering
\includegraphics[width=0.99\linewidth]{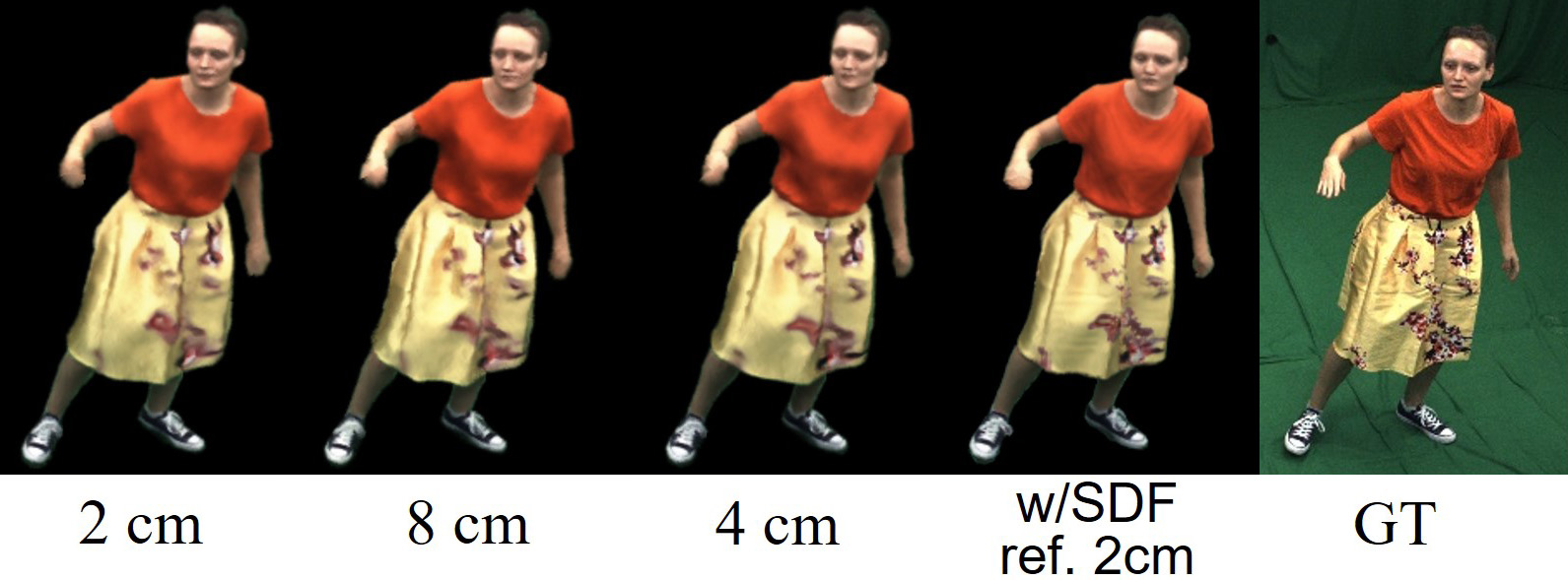}
\caption{
\textbf{Height of UTTS.}
The rendering generated with different configurations on the height of UTTS.
The results demonstrate that setting the UTTS height to \textbf{4 cm} in the Field Pre-Training stage outperforms the alternative settings (\textbf{2 cm} and \textbf{4 cm}). Moreover, after refining the underlying template mesh with the implicit field, i.e., SDF-driven Surface Refinement stage, we can shrink the height $d_\mathrm{max}$ of the UTTS space to 2 cm (\textbf{w/SDF ref. 2cm}), and achieve even better accuracy.
}
\label{fig:supplablationheight}
\end{figure}

As mentioned in Sec.~\ref{sec:supplmap}, the height $d_\mathrm{max}$ of the Undeformed Tri-plane Texture Space (UTTS) significantly impacts the mapping ambiguity of the spatial samples. 
To this end, we conducted an ablation study that assesses different settings on the height of UTTS for the field pre-training stage. 
As illustrated in Tab.~\ref{tab:supplheight} and Fig.~\ref{fig:supplablationheight}, smaller UTTS height (termed as \textbf{2 cm}) reduces mapping collisions but may miss the "real surface" during field pre-training, especially for loose clothing. 
Conversely, a larger UTTS height (termed as \textbf{8 cm}) introduces more collisions. 
The final design choice for the UTTS height in the field-pretraining stage (termed as \textbf{4 cm}) outperforms the alternative settings in both view synthesis and shape reconstruction.
\par 
After the SDF-driven surface refinement, with the explicit surface closer to the "real surface," we can shrink the height of the UTTS to 2cm (termed as \textbf{w/SDF ref. 2cm}) to achieve even better accuracy due to further reduced collisions.
\begin{table}[h]
\renewcommand\tabcolsep{10.0pt}
\small
    \centering
    \caption{\textbf{Height of UTTS}.
    Quantitative evaluation of various settings regarding the height of UTTS during the field-pretraining stage. Note our final design choice, i.e., setting the height of the UTTS as \textbf{4 cm} in the field-pretraining stage yields the highest accuracy in both view synthesis and shape reconstruction. Notably, our full model, denoted as \textbf{w/SDF ref. 2cm}, further improves the accuracy in view and geometry synthesis.
    }
    \label{tab:supplheight}
    \begin{tabular}{l|c|c|c}
    \hline
    \multicolumn{4}{c}{\textit{Training Poses (Loose Clothing)}} \\
    \hline
    Methods & \textbf{PSNR} $\uparrow$  & \textbf{LPIPS} $\downarrow$ & \textbf{Cham.} $\downarrow$ \\
    \hline
    2 cm      & 30.52  & 23.51 & 1.877  \\
    8 cm       & 30.36  & 23.20 & 1.783 \\
    \textbf{4 cm}      & \textbf{30.55}  & \textbf{22.96} & 
    \textbf{1.714} \\
    \hline
    \textbf{w/SDF ref. 2cm}       & \textbf{31.68}  & \textbf{16.14} & \textbf{1.488}  \\
    
    \hline
    \end{tabular}
\end{table}

%
%
%
\section{TriHuman Viewer} \label{supplsec:system}
To demonstrate the full potential of TriHuman, we proposed TriHuman Viewer --- a real-time, interactive system for visualizing and editing high-quality clothed human avatars with various motions. 
In this section, we explain into the TriHuman Viewer from the following perspectives: user interface (Sec.~\ref{supplsubsec:userinterface}), supported interaction (Sec. ~\ref{supplsubsec:functions}), and the runtime analysis (Sec. ~\ref{supplsubsec:runtime}) for each component. 
For a more comprehensive visualization of our system, please refer to the supplementary video.
%
%
\subsection{User Interface} \label{supplsubsec:userinterface}
Fig. ~\ref{supplsubsec:userinterface} illustrates the user interface of our TriHuman Viewer. 
The user interface of TriHuman Viewer can be deployed on a personal laptop computer, which consists of three main components, i.e., the control panel (Fig. ~\ref{supplsubsec:userinterface}(A)), the character view (Fig. ~\ref{supplsubsec:userinterface}(B)), and the render view (Fig. ~\ref{supplsubsec:userinterface}(C)).
The control panel contains the settings for neural rendering character visualization, e.g., static-viewpoint rendering or free-viewpoint rendering mode, character viewing or editing mode. 
Users may also select the desired frame to be visualized from the training/testing motions. 
The character view is a 3D viewer for visualizing the skeletal pose and the generated detailed geometry. 
Moreover, the render view shows the imagery rendered with the TriHuman backend model, which is deployed on a server, using the current skeletal motion and camera poses. 

\begin{figure}[h]
\includegraphics[width=0.98\linewidth]{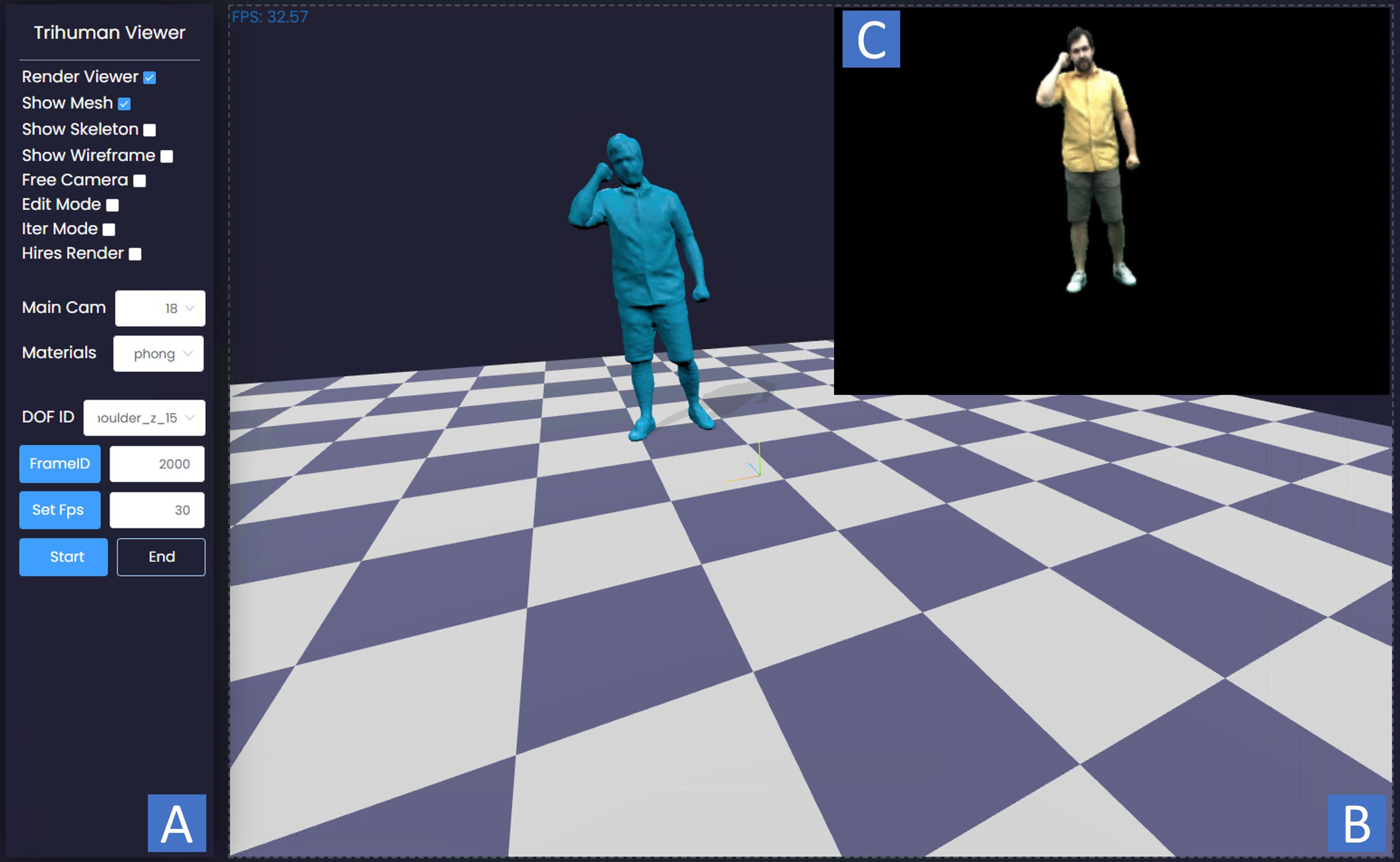}
  \caption{
    The interface of TriHuman Viewer. The TriHuman Viewer is adopted to view and create high-fidelity, clothed human imagery and geometry in real time. 
    The user interface of the Trihuman consists of three sub-views, namely, the (A) control panel, used for configuring the rendering and geometry generation modes and parameters; the (B) character view, adopted for inspecting the skeletal motion and the geometry of the clothed human; (C) the rendering view, presenting the neural-rendered results from TriHuman backend model. 
  }
  \label{fig:supplinterface}
\end{figure}
%
%
\subsection{Supported Functionalities} \label{supplsubsec:functions}
The major functionalities supported by TriHuman Viewer can be summarized as follows: real-time instant replay, DOF editing, and free-viewpoint rendering.
\par
\noindent\textbf{Real-time Instant Replay.} 
The TriHuman Viewer is set to real-time instant replay mode by default, where users may browse the current skeletal pose, detailed geometry, and the rendering from the selected studio camera. 
The geometry and rendering are generated in real time with the TriHuman model deployed on the server and are sent to the TriHuman Viewer front-end with a WebSocket. 
In the runtime analysis section, we elaborate on the runtime for each component in more detail.
\par
\noindent\textbf{DOF Editing.}
Apart from real-time instant replay, the TriHuman Viewer also supports creating high-fidelity clothed human geometry and rendering with novel poses, thanks to TriHuman's generalization ability to novel poses. 
In the DOF editing mode, users may create novel skeletal motions by modifying the character's skeleton DOFs. 
Meanwhile, the TriHuman backend model automatically produces the corresponding clothed human meshes and renderings. 
Specifically, to ease the complexity for novice users to create plausible skeletal motion input, we allow users to edit the skeletal poses starting from any training or test frames.
\par
\noindent\textbf{Free-Viewpoint Rendering.} 
The camera calibration for the TriHuman Viewer is configured with the studio camera by default.
However, users have the option to switch to Free-viewpoint rendering mode. 
In this mode, users can inspect the skeletal poses and the generated geometry with a panoramic camera orbiting around the character.
Concurrently, the rendering view presents the clothed character's imagery with the camera pose synchronized with the character view, computed with the TriHuman backend models deployed on the server.
%
%
\subsection{Runtime Analysis} \label{supplsubsec:runtime}
The main paper highlights the TriHuman Viewer's capability to deliver a real-time experience in producing high-fidelity geometry and rendering with a merely one-frame latency.
In the following section, we delve into the runtime of the TriHuman backend concerning the generation of geometry and images.
In practice, we can achieve geometry generation and rendering at more than 25 frames per second with a one-frame delay.
\par
\noindent\textbf{Real-time Geometry Generation.} 
The real-time geometry generation component takes the skeletal motion as input and outputs the consistent geometry with around $30{,}000$ vertices. 
The whole component consists of two pivotal stages: explicit asset preparation and implicit geometry generation. 
\par
The explicit assets preparation computes the clothed human mesh from a sliding window of previous skeletal poses.
Additionally, it renders motion texture maps from the clothed human mesh. 
In practice, the generation of the clothed human mesh requires approximately \textbf{25 milliseconds}, while rendering motion-aware feature maps takes \textbf{less than 1 millisecond}. 
Specifically, the explicit assets preparation runs within a background thread, and the generated outputs are then piped to the implicit geometry generation.
\par
The implicit geometry generation runs as the foreground thread, which first produces a signed distance field (SDF) upon the UTTS space of the clothed human mesh. 
Subsequently, the clothed human template mesh generated from the explicit module is deformed guided by the SDF. 
To be more detailed, the implicit geometry generation comprises the following main steps. 
\begin{itemize}
    \item Generating motion-aware triplanes with the triplane generator takes approximately \textbf{4 milliseconds}. 
    \item Mapping the template mesh vertices from the observation space to the UTTS space requires about \textbf{8 milliseconds}. 
    \item Sample features from the motion-aware triplanes, which usually take \textbf{2 milliseconds}.
    \item Computing the SDF value from the sampled triplane features, which takes approximately \textbf{6 milliseconds}. 
    \item Determining the moving directions for the coarse-clothed human mesh through gradient computation in the SDF, which takes around \textbf{6 milliseconds}.
\end{itemize}
\par 
In practice, we opt to distribute the tasks of implicit and explicit generation threads across two Nvidia A100 graphic cards. 
\par 
\noindent\textbf{Real-time Image Generation.}
The real-time image generation component takes the skeletal motion and the virtual camera view as inputs and produces photorealistic rendering at a resolution of 0.5K. 
Similar to the geometry generation pipeline, we divide the image generation pipeline into two sub-tasks, i.e., the explicit assets preparation runs in the background thread, and the implicit-based rendering takes place in the foreground thread. 
Each thread is running on a different Nvidia A100 graphics card.
\par
In addition to creating the clothed human template and rendering motion-aware feature maps, the explicit assets preparation step for image generation also involves rendering the character's depth map from the active camera in TriHuman Viewer. 
The generated depth map is adopted for filtering out the off-the-surface rays before ray marching in the implicit-based rendering module, taking approximately \textbf{10 milliseconds}.
The implicit rendering module running in the foreground consists of the following major steps: 
\begin{itemize}
    \item Generating motion-aware triplanes, which require around \textbf{4 milliseconds}.
    \item Mapping ray samples from the observation space to the UTTS space, which takes approximately \textbf{10 milliseconds}. In practice, we adopt 20 samples for each foreground ray for ray-marching-based neural rendering, striking a nice balance between the rendering quality and the execution speed. Moreover, the samples that fall out of the range of UTTS are excluded from later computation, further speeding up the evaluation process. 
    \item Performing the forward pass in SDF to compute position-aware features for the ray samples in the UTTS space, with a duration of around \textbf{8 milliseconds}.
    \item Sample features from the motion-aware triplanes, which usually take \textbf{2 milliseconds}.
    \item Executing the backward pass in SDF to compute the normal values for the ray samples in the observation space, which also takes approximately \textbf{8 milliseconds}.
    \item Conducting the forward pass in the color network to compute the color for each ray sample, consuming around \textbf{7 milliseconds}.
\end{itemize}

\end{document}